\PassOptionsToPackage{table}{xcolor} 
\documentclass{article}

\PassOptionsToPackage{numbers, compress}{natbib}
\usepackage[preprint]{neurips_2024}

\usepackage[utf8]{inputenc} 
\usepackage[T1]{fontenc}    
\usepackage{hyperref}       
\usepackage{url}            
\usepackage{booktabs}       
\usepackage{amsfonts}       
\usepackage{nicefrac}       
\usepackage{microtype}      
\usepackage{xcolor}         
\usepackage{enumitem} 
\usepackage{graphicx}
\usepackage{listings}
\usepackage{floatflt}
\usepackage{wrapfig}
\usepackage{amsmath}
\usepackage{multirow}
\usepackage{subcaption}
\usepackage[most]{tcolorbox}
\usepackage{inconsolata}
\usepackage{float}
\usepackage{longtable}  
\usepackage{array}      
\usepackage[utf8]{inputenc} 

\title{AutoSOTA: An End-to-End Automated Research System for State-of-the-Art AI Model Discovery}
    
\author{
\textbf{Yu Li}$^{1,*}$\hspace{3mm}
\textbf{Chenyang Shao}$^{1,2,*}$\hspace{3mm}
\textbf{Xinyang Liu}$^{1,*}$\hspace{3mm} 
\textbf{Ruotong Zhao}$^1$\hspace{3mm} 
\textbf{Peijie Liu}$^1$\hspace{3mm} \\
\textbf{Hongyuan Su}$^{1,2}$\hspace{3mm}
\textbf{Zhibin Chen}$^1$\hspace{3mm} 
\textbf{Qinglong Yang}$^1$\hspace{3mm}
\textbf{Anjie Xu}$^{2,3}$\hspace{3mm} 
\textbf{Yi Fang}$^{2,4}$\hspace{3mm} 
\textbf{Qingbin Zeng}$^1$\hspace{3mm}  \\
\textbf{Tianxing Li}$^1$\hspace{3mm}  
\textbf{Jingbo Xu}$^1$\hspace{3mm}
\textbf{Fengli Xu}$^{1,2,\dagger}$\hspace{3mm} 
\textbf{Yong Li}$^{1,2,\dagger}$\hspace{3mm}
\textbf{Tie-Yan Liu}$^{2}$\hspace{3mm}
\\
$^1$Department of Electronic Engineering, BNRist, Tsinghua University\\
$^2$Zhongguancun Academy\\
$^3$Peking University\\
$^4$University of Science and Technology of China\\
$*$\textit{Equal contribution.}\\
$\dagger$\{fenglixu, liyong07\}@tsinghua.edu.cn
\vspace{1mm}\\
\parbox{0.1\textwidth}
{\includegraphics[width=0.4\linewidth]{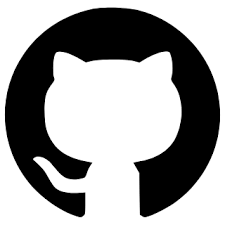}}\hspace{-7mm}{\nolinkurl{https://tsinghua-fib-lab.github.io/AutoSOTA/}}\vspace{-3mm}
}

\begin{document}

\maketitle
\begin{abstract}
Artificial intelligence research increasingly depends on prolonged cycles of reproduction, debugging, and iterative refinement to achieve State-Of-The-Art (SOTA) performance, creating a growing need for systems that can accelerate the full pipeline of empirical model optimization. In this work, we introduce AutoSOTA, an end-to-end automated research system that advances the latest SOTA models published in top-tier AI papers to reproducible and empirically improved new SOTA models. We formulate this problem through three tightly coupled stages: resource preparation and goal setting; experiment evaluation; and reflection and ideation. To tackle this problem, AutoSOTA adopts a multi-agent architecture with eight specialized agents that collaboratively ground papers to code and dependencies, initialize and repair execution environments, track long-horizon experiments, generate and schedule optimization ideas, and supervise validity to avoid spurious gains. We evaluate AutoSOTA on the recent research papers collected from eight top-tier AI conferences under filters for methodological relevance, code availability, repository readiness, and execution cost. Across these papers, AutoSOTA achieves strong end-to-end performance in both automated replication and subsequent optimization. Specifically, it successfully discovers 105 new SOTA models that surpass the original reported methods, averaging approximately five hours per paper. Case studies spanning LLM, NLP, computer vision, time series, and optimization further show that the system can move beyond routine hyperparameter tuning to identify architectural innovation, algorithmic redesigns, and workflow-level improvements. These results suggest that end-to-end research automation can serve not only as a performance optimizer, but also as a new form of research infrastructure that reduces repetitive experimental burden. Ultimately, AutoSOTA points toward a fundamental shift in the research paradigm, fostering a tightly coupled human-AI collaboration. By delegating large-scale execution, reproduction, and reflection to automated agents, this framework empowers human scientists to redirect their attention toward deeper conceptual innovation and disruptive questions, sharply returning human effort to the irreplaceable core of scientific discovery.
\end{abstract}

\newpage

\begingroup
\renewcommand{\baselinestretch}{1}\normalsize
\tableofcontents
\endgroup

\newpage

\section{Introduction}

Artificial intelligence research is increasingly organized around the pursuit of State-Of-The-Art (SOTA) performance. In many areas, the value of a new method is judged not only by conceptual novelty, but by whether it can surpass a highly competitive empirical frontier. Yet, achieving such SOTA breakthroughs is almost never the product of a standalone theoretical insight. Instead, the fundamental bottleneck lies in a prolonged cycle of implementation, replication, debugging, evaluation, and reflection, in which months of human effort are spent translating promising insights into reproducible performance gains. This tension is severely exacerbated in modern AI, where research artifacts are distributed across papers, repositories, datasets, checkpoints, and undocumented engineering assumptions. As a result, a substantial portion of scientific attention is absorbed by repetitive experimental iteration, which might otherwise be directed toward more original, disruptive exploration. To alleviate this burden, our previous work introduced OmniScientist\cite{shao2025omniscientist}, a comprehensive research agent that pushes the paradigm toward fully autonomous, end-to-end scientific discovery. Building upon this overarching framework, an important subsequent challenge is whether we can instantiate this capability into targeted automated systems that reliably accelerate the highly competitive pipeline of SOTA model optimization. 

Recent progress suggests that this goal is increasingly plausible, but existing efforts remain incomplete. Traditional automated machine learning (AutoML)\cite{barbudo2023eight,he2021automl,karmaker2021automl,li2025agentexpt} systems have demonstrated the value of hyperparameter tuning and architecture search, while newer LLM-based systems have shown impressive capabilities in code generation, debugging, and restricted forms of algorithm discovery. At the same time, emerging autonomous AI scientist systems have begun to explore broader research workflows. However, a critical limitation persists: existing frameworks are severely constrained by narrow problem formulations. For instance, even the latest iterative frameworks like AutoResearch\cite{karpathy2026autoresearch} strictly confine the agent to improving a single bounded code module within a sanitized sandbox, entirely sidestepping the complexities of environment setup. They structurally fail to address the full end-to-end problem of starting from an unstructured research paper, recovering the corresponding executable environment, reproducing the reported baseline, and then performing valid, open-ended methodological reflection under realistic constraints. Overcoming this precise gap constitutes the primary motivation for our work: developing an end-to-end automated AI research system that can bridge unstructured literature, experiment environments setup, and high-level reflection in a single closed-loop framework.

In this work, we formulate this challenge as an end-to-end discovery problem that maps a top-tier AI paper to an improved executable repository whose empirical performance surpasses the original method. To make this open-ended task tractable, we conceptually structure the core technical challenge around three tightly coupled sub-problems. The first fundamental hurdle is resource preparation and goal setting, which requires grounding a paper into a concrete experimental task by locating repositories, datasets, base models, and the exact target metric to be exceeded. The second major obstacle involves experiment evaluation, which requires transforming incomplete and noisy academic code into a runnable baseline that faithfully approximates the reported results despite missing scripts, broken dependencies, and long-horizon debugging challenges. Finally, the most demanding phase is reflection and ideation, which elevates the system from a mere replication engine to an autonomous scientific discoverer. At this stage, the system is tasked with uncovering genuine algorithmic improvements—exploring structural innovations, novel learning objectives, and optimized implementation strategies—while intelligently balancing open-ended exploration against the exorbitant computational cost of real-world trials. Together, these three sub-problems define the core scientific and engineering bottlenecks that any end-to-end automated research system must solve. 

To address these challenges, we introduce AutoSOTA, a multi-agent framework designed to mimic the division of labor in human AI research. Rather than executing a monolithic routine, AutoSOTA strategically decomposes the workflow across eight specialized agents organized into the three stages above. Addressing the initial preparation bottleneck, AgentResource grounds research papers to repositories and external dependencies, while AgentObjective constructs a tree-structured evaluation rubric that translates the paper’s macro goal into dense, verifiable feedback. To overcome execution fragility, AgentInit builds the executable environment, AgentMonitor tracks long-horizon execution and prevents deadlock, and AgentFix performs timely repair of runtime failures. Driving the actual scientific discovery, AgentIdeator constructs a constrained hypothesis library, AgentScheduler manages iterative optimization across code states and compute resources, and AgentSupervisor enforces a red-line system that prevents invalid metric gains caused by evaluation leakage, protocol violations, or other shortcuts. Through this highly orchestrated modular design, AutoSOTA converts fragmented academic artifacts into an execution-ready optimization loop and supports stable, scalable, and scientifically comparable SOTA-oriented research automation.

We evaluate AutoSOTA on the research papers drawn from eight top-tier AI conferences, after filtering for methodological relevance, public code availability, repository readiness, and tractable execution cost. Across these curated research papers, AutoSOTA definitively proves its efficacy by demonstrating strong end-to-end performance in both automated replication and subsequent optimization, successfully discovering new SOTA models for 105 papers. Remarkably, the system yields an average performance improvement of nearly 10\% over the original main methods, requiring only about five hours of execution time per paper. We further presents case studies spanning research subfields of LLM, NLP, computer vision, time series, and optimization, illustrating that the framework has the profound capability to transcend routine hyperparameter tuning and execute substantive architectural, algorithmic, and workflow-level interventions. These results suggest that end-to-end research automation is no longer limited to narrow coding tasks, but can support broad empirical advancement across heterogeneous AI research areas. 

More broadly, the overarching significance of AutoSOTA lies well beyond raw benchmark improvement. As described in the project vision, the system is intended not merely as an optimizer, but as a fundamentally new research infrastructure for AI science itself: one that can absorb the tedious, high-frequency cycle of experimental iteration and thereby amplify human creativity. In this sense, AutoSOTA points toward a shift in the research paradigm. Instead of treating scientific automation as a tool for isolated assistance, it suggests the possibility of tightly coupled human–AI collaboration in which automated agents handle large-scale execution, reproduction, and reflection, while human scientists redirect their attention toward deeper conceptual innovation and more disruptive questions. Such a direction does not diminish the role of human researchers. Rather, it sharpens it by moving human effort away from repetitive optimization and back toward the irreplaceable core of scientific discovery.

\section{Research Problem} 
\label{sec:problem}

\begin{figure}[htbp]
    \centering
    \includegraphics[width=\linewidth]{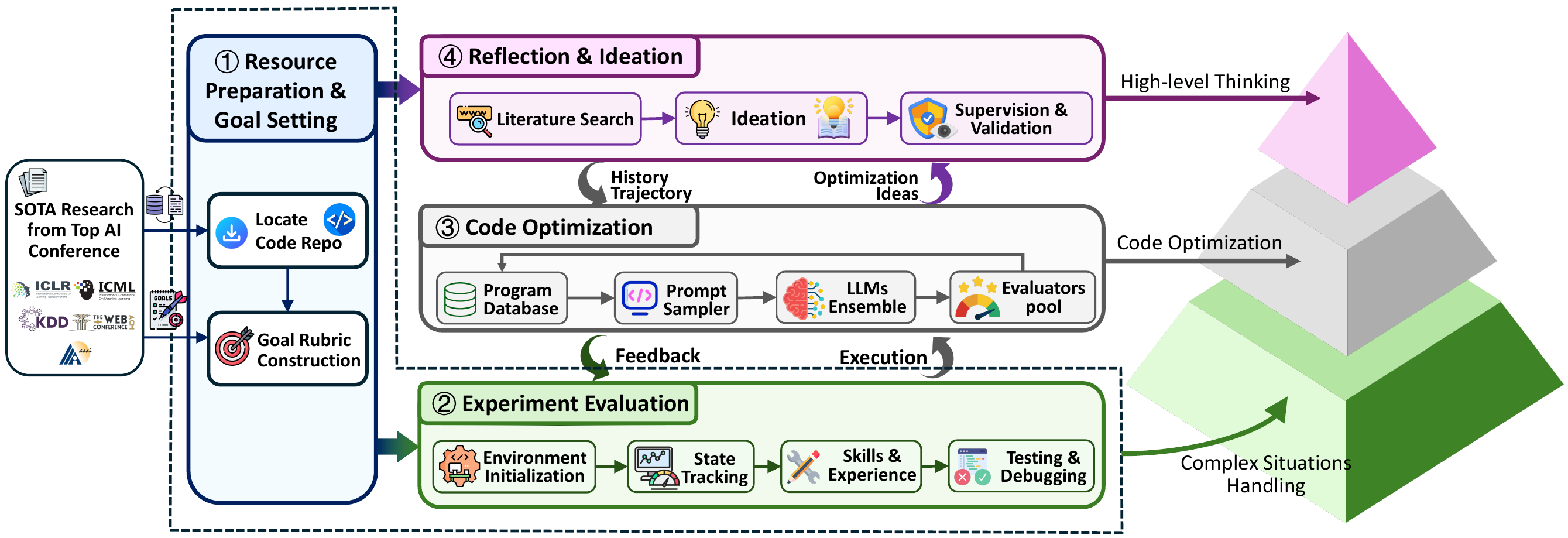}
    \caption{Task Definition of AutoSOTA: An End-to-End Automated Research Framework}
    \label{fig:research_problem}
\end{figure}

The overarching objective of AutoSOTA is to develop an end-to-end automated research system for SOTA model discovery. While systems like Alphaevolve~\cite{novikov2025alphaevolve} have made significant strides in pure code optimization, AutoSOTA advances this paradigm by autonomously converting unstructured academic literature into empirically superior algorithms that surpass the published baselines. Formally, we define this end-to-end task as an open-ended discovery function $\mathcal{F}: \mathcal{P} \rightarrow \mathcal{R}^{*}$, where the input $\mathcal{P}$ represents an unstructured, multi-modal research paper from a top-tier AI conference, and the output $\mathcal{R}^{*}$ denotes an advanced, executable GitHub-style repository containing a novel algorithmic variant that strictly outperforms the method proposed in $\mathcal{P}$. Unlike standard code generation tasks that operate within well-defined programmatic constraints, AutoSOTA navigates the highly stochastic, exploratory domain of scientific innovation. To render this massive undertaking tractable, we explicitly decompose the global objective into the following four sequential, tightly coupled sub-problems, each presenting unique, highly non-trivial challenges that restrict the capabilities of current LLM agents:

\textbf{Resource Preparation and Goal Setting:} This initial phase is tasked with grounding the theoretical paper into a concrete experimental context to establish a baseline for future innovation. We formalize this as $f_{prep}: (\mathcal{P}, \mathcal{W}) \rightarrow (\mathcal{C}, \mathcal{D}, \mathcal{M}, g^{*})$, where the system must interact with the open web $\mathcal{W}$ to extract the foundational codebase $\mathcal{C}$, necessary datasets $\mathcal{D}$, base models $\mathcal{M}$, and precisely define the target performance metric $g^{*}$. The primary bottleneck here is the severe fragmentation and implicit coupling of academic resources. Repository links are rarely standardized, often buried in footnotes or external project pages rather than abstracts, and critical dependencies are frequently dispersed across undocumented GitHub issues or disconnected supplementary materials. Furthermore, establishing the baseline replication goal $g^{*}$ is extraordinarily difficult due to the chaotic presentation of empirical results in literature. Papers typically feature dense, multi-dimensional tables with convoluted headers and varying baselines. An autonomous agent must possess deep semantic reasoning to disentangle multi-metric trade-offs and accurately pinpoint the single most scientifically meaningful performance number to serve as the rigorous benchmark it must eventually defeat.

\textbf{Experiment Evaluation:} Following the acquisition of resources and objectives, the system must undertake this evaluation phase, defined as synthesizing a functional repository state $\mathcal{R}_{rep}$ such that its empirical evaluation closely approximates the target metric, $\text{eval}(\mathcal{R}_{rep}) \approx g^{*}$. This phase transforms the static resources into a dynamic execution pipeline, serving as the springboard for algorithmic evolution. The fundamental difficulty of this sub-task lies in its nature as an extreme long-horizon reasoning and acting problem under partial observability. The agent is forced to navigate a labyrinth of environment configurations, implicit hardware dependencies, and deprecated library versions. During this protracted execution phase, agents frequently succumb to local optima and algorithmic dead-ends, most notably entering infinite loops of hallucinated bug-fixing during environment setup. Moreover, open-source academic code is notoriously incomplete, often plagued by missing utility scripts, hardcoded absolute local paths, or undocumented pre-processing steps. The agent must not only diagnose these silent failures and cryptic tracebacks but also synthesize the missing logic from the conceptual descriptions in the paper, bridging the perilous gap between abstract mathematics and concrete engineering.

\textbf{Code Optimization:} Acting as the operational bridge between baseline reproduction and high-level scientific innovation, this intermediate phase is responsible for executing concrete programmatic modifications. We formalize this process as $f_{opt}: (\mathcal{R}_{rep}, \mathcal{I}) \rightarrow \mathcal{R}_{cand}$, where the system translates abstract optimization ideas ($\mathcal{I}$) into executable candidate repositories ($\mathcal{R}_{cand}$). The primary challenge lies in accurately mapping structural or algorithmic concepts into syntactically correct, context-aware code edits within a complex, multi-file repository. The system must orchestrate program databases, prompt samplers, and LLM ensembles to efficiently generate code variations, while utilizing an evaluator pool to gather rapid execution feedback.

\textbf{Reflection \& Ideation:} As the overarching cognitive engine and the final frontier of the AutoSOTA pipeline, this phase elevates the system from a code-editing engine to an autonomous scientific discoverer. This sub-task is formalized as an open-ended search and validation process $f_{refine}: \mathcal{R}_{cand} \rightarrow \mathcal{R}^{*}$ subject to the strict constraint that $\text{eval}(\mathcal{R}^{*}) > \text{eval}(\mathcal{R}_{rep})$. By outsourcing low-level syntax modifications to the code optimization phase, the agent is freed to engage in high-level thinking: performing literature search, generating novel hypotheses, and enforcing rigorous supervision. The agent must iteratively reflect on historical execution trajectories and propose profound structural innovations to network architectures or loss functions, rather than defaulting to trivial hyperparameter tuning. The predominant challenge in this expansive search space is the severe imbalance between exploration and exploitation, heavily compounded by the exorbitant computational cost of evaluating a single experimental trial. Consequently, the core research problem lies in designing a mechanism that can efficiently synthesize historical experimental feedback, systematically avoid cyclical reasoning, and ensure that all metric gains are methodologically valid.

Ultimately, realizing AutoSOTA requires overcoming the compounding errors that propagate through these four stages. The transition from unstructured knowledge extraction to long-horizon software engineering, iterative code mutation, and finally to autonomous scientific discovery, demands a robust, fault-tolerant framework capable of navigating the immense ambiguity inherent in pushing the boundaries of cutting-edge AI research.






\section{AutoSOTA Framework}
\subsection{Architecture Design}
\label{subsec:architecture}

\begin{figure}[htbp]
    \centering
    \includegraphics[width=\linewidth]{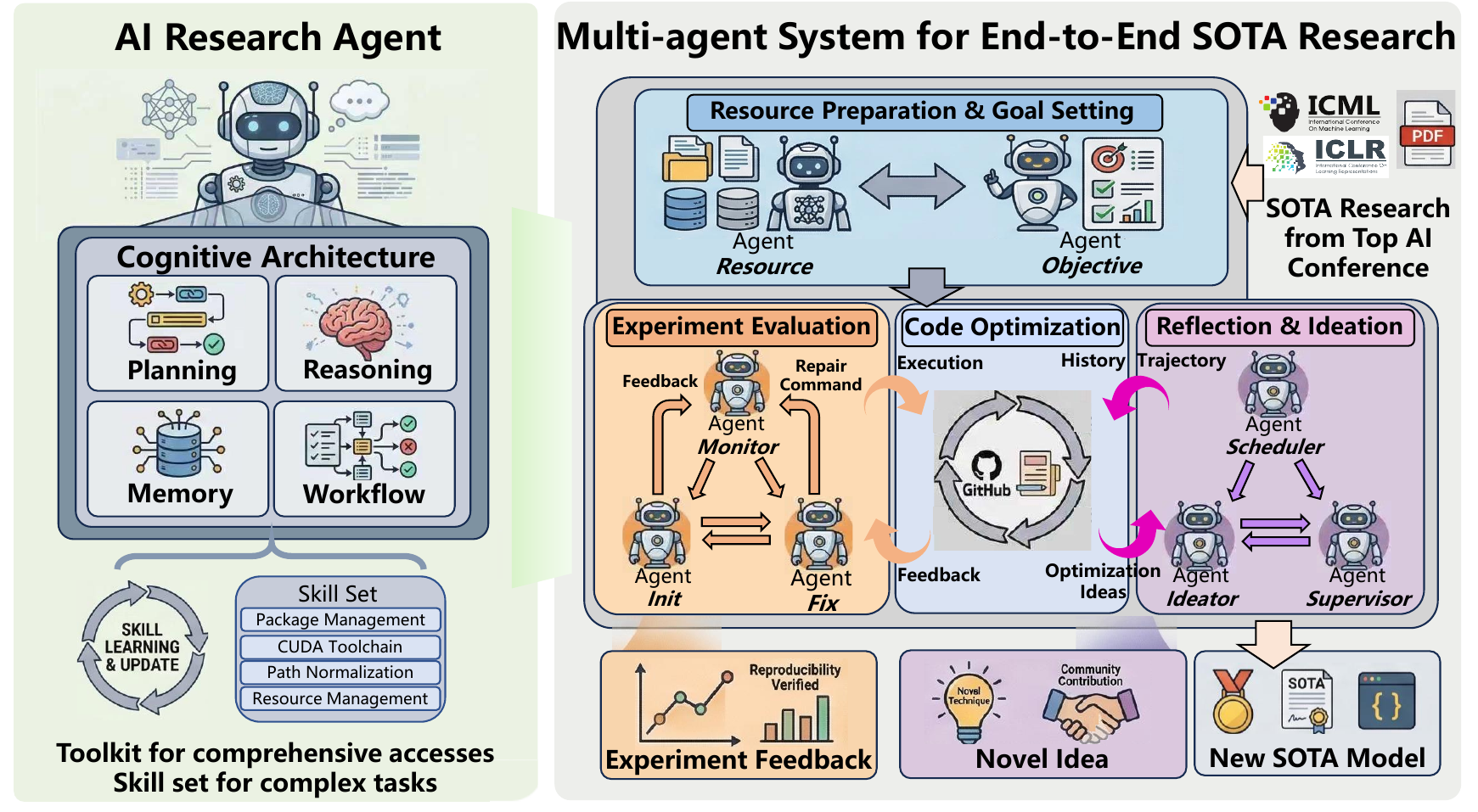}
    \caption{Overall Framework of AutoSOTA}
    \label{fig:Overall_framework}
\end{figure}

To realize the end-to-end discovery function $\mathcal{F}$ defined in Section \ref{sec:problem}, we architect AutoSOTA as a multi-agent system that mimics the workflow of human scientists. To avoid the inevitable collapse of a monolithic LLM prompt under the extreme context lengths and compounding errors of long-horizon scientific execution, we systematically decompose the cognitive and operational workload across eight specialized, strictly bounded agents. While the mechanical generation of syntax modifications---often framed as \textbf{Code Optimization}---has been the primary focus of recent evolutionary frameworks like AlphaEvolve~\cite{novikov2025alphaevolve}, we treat this programmatic translation as a standard operational bridge and do not explicitly focus on it here. Instead, our architecture is strategically orchestrated around three core macro-modules: Resource Preparation and Goal Setting, Experiment Evaluation, and Reflection \& Ideation (see Fig.~\ref{fig:Overall_framework}). Crucially, rather than functioning as a strictly sequential pipeline, Experiment Evaluation and Reflection \& Ideation operate in a tightly coupled, iterative cycle to continuously drive algorithmic evolution.

The overarching workflow commences with the \textbf{Resource Preparation and Goal Setting} module, which is designed to bridge the chasm between unstructured academic literature and actionable engineering constraints. This foundational stage is governed by two complementary agents. \textbf{AgentResource} is responsible for the physical grounding of the task, acquiring the target code repositories associated with raw conference papers ($\mathcal{P}$) while autonomously discovering and downloading heavyweight external dependencies (e.g., datasets and base models). Simultaneously, \textbf{AgentObjective} addresses the diagnostic opacity of automated replication by parsing the multi-modal academic context to construct a tree-structured, dense evaluation rubric. Together, these two agents ensure that downstream execution is anchored to a concrete codebase and a mathematically rigorous target metric ($g^{*}$).

Once the static assets and evaluative objectives are materialized, the system enters the core optimization loop, fundamentally driven by \textbf{Experiment Evaluation}. Transforming a noisy, historically brittle academic repository into a faithful, executable baseline ($\mathcal{R}_{rep}$)---and subsequently evaluating new algorithmic variants---requires robust operational resilience. We distribute this dynamic execution responsibility across a triad of interactive agents. \textbf{AgentInit} serves as the workflow-grounded pioneer, utilizing a library of preset skills to construct the initial execution environment and synthesize missing repository logic. Because autonomous execution is prone to pathological dead-ends, \textbf{AgentMonitor} acts as an external, real-time supervisor, tracking the execution trace to prevent infinite debugging loops and enforce global budgets. When execution inevitably encounters systematic errors, \textbf{AgentFix} steps in as the skill-augmented repair module, leveraging a cross-task failure memory to inject protocol-preserving engineering fixes. 

Operating in continuous synergy with the evaluation phase is the \textbf{Reflection \& Ideation} module, which elevates the system from a sophisticated code-replication engine to an autonomous scientific discoverer. To successfully navigate the expansive algorithmic search space and achieve the optimized state $\mathcal{R}^{*}$, the architecture employs three highly coupled agents. \textbf{AgentIdeator} initiates the reasoning process not through unconstrained brainstorming, but by constructing a structured, protocol-respecting hypothesis library grounded in domain prior knowledge. \textbf{AgentScheduler} acts as the central orchestration core for the iterative loop, seamlessly managing dynamic resource allocation, version control, and external memory contexts over multi-day experiments, continuously routing newly generated ideas back into the evaluation phase for physical execution. Finally, to prevent the system from achieving superficial metric gains through invalid shortcuts, \textbf{AgentSupervisor} acts as the strict guardian of scientific integrity, enforcing a non-negotiable ``Red Line System'' to guarantee that all algorithmic improvements remain strictly comparable to the original baseline.

Ultimately, this eight-agent architecture forms a cohesive, fault-tolerant discovery engine. AutoSOTA explicitly bounds the responsibilities of each module by separating physical resource acquisition from logical evaluation, execution from external supervision, and hypothesis generation from scientific validation. This strict architectural decoupling, combined with the continuous iterative feedback loop between execution and reflection, systematically mitigates the compounding ambiguities inherent in cutting-edge AI research, enabling the stable, autonomous advancement of state-of-the-art methodologies.

\subsection{AgentResource: Bridging Static Literature and Executable Environments}

\begin{figure}
    \centering
    \includegraphics[width=\linewidth]{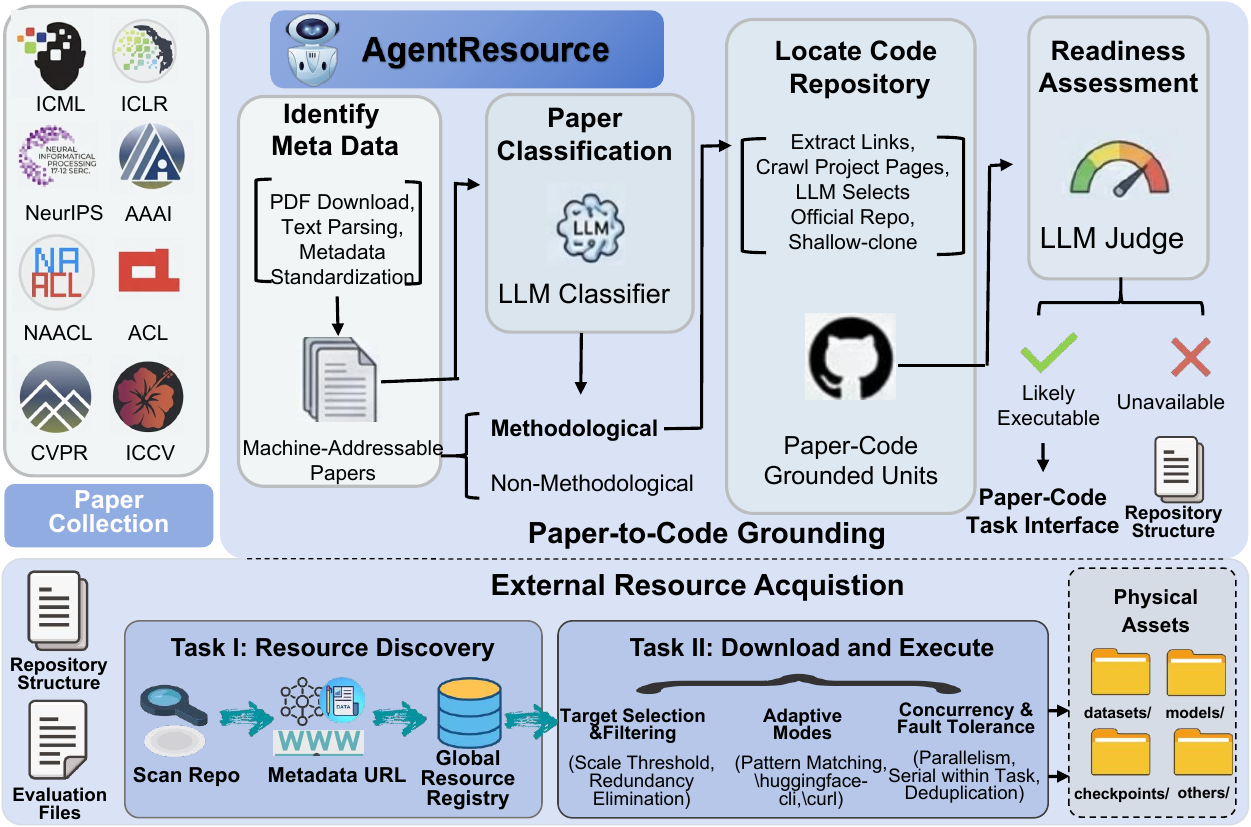}
    \caption{Overall Framework of the Resource Acquisition Process}
    \label{fig:Resource_framework}
\end{figure}

To enable automated optimization and evaluation, the system must first bridge the gap between static, human-readable publications and fully executable experimental environments. As illustrated in Figure~\ref{fig:Resource_framework}, this complex transformation is managed by the \textbf{AgentResource} module through a unified, two-stage pipeline. First, the system performs paper-to-repository grounding to identify, extract, and normalize the foundational code artifacts associated with candidate papers. However, because modern AI repositories rarely self-contain their requisite massive datasets or pre-trained weights, a subsequent external resource acquisition phase is employed to automatically discover and physically download these critical dependencies. Together, these two sub-modules seamlessly convert raw academic literature into fully self-contained, execution-ready task units for downstream processing.

\subsubsection{Paper-to-Repository Grounding and Curation}
Scientific papers are written for human readers rather than direct machine execution. The artifacts needed for replication are often incomplete, distributed across multiple sources, or only mentioned implicitly in PDFs, project pages, repositories, and auxiliary files. This makes it difficult to start large-scale scientific optimization directly from raw conference papers. Before downstream modules can initialize experiments or allocate execution resources, the system must identify which papers correspond to executable empirical tasks, locate their code artifacts, and determine whether those artifacts are usable for further processing. We refer to this stage as \textbf{AgentResource}, which addresses the problem of grounding papers to repositories.

AgentResource starts from conference-level metadata and converts raw paper collections into a unified local representation with standardized metadata, downloaded PDFs, and parsed text. It operates on papers from eight top-tier conferences in 2025, including ICML, ICLR, NeurIPS, AAAI, NAACL, ICCV, ACL, and CVPR. This step transforms heterogeneous conference outputs into machine-addressable paper units that can be processed by later modules. Given this normalized paper pool, AgentResource next performs paper typing to identify candidates for empirical optimization. It uses an LLM-based classifier to separate methodological papers from other categories, such as theoretical, benchmark, analysis, and position papers. The classifier is prompted to judge whether the main contribution is a new method, model, algorithm, framework, or system that is experimentally validated against baselines. This step narrows an open-domain literature pool into a set of candidate tasks with experimentally grounded objectives.

For papers that pass this filter, AgentResource then carries out repository discovery and normalization. It first extracts GitHub links from the front matter of paper PDFs, covering both direct repository links and indirect project pages such as \texttt{github.io} sites. When a project page is found, the system crawls that page to recover the underlying repository link. If multiple candidate repositories are associated with one paper, AgentResource uses an LLM selector to identify the most relevant official repository, while filtering template repositories and removing duplicates. The selected repository is then shallow-cloned into a normalized local directory structure indexed by conference, paper ID, and repository owner. To improve recall, the workflow also performs a full-document scan when repository links are not found in the initial pass. After repository grounding, AgentResource performs a lightweight readiness assessment before more expensive execution begins downstream. For each cloned repository, the system collects three compact signals: the paper abstract, a shallow file tree of the repository, and the opening portion of the README. An LLM judge then estimates whether the repository is likely actionable for reproducing the main result, using a deliberately permissive criterion that excludes only clear placeholders or repositories that explicitly state that the code has not been released. 

Therefore, the output of AgentResource is a structured task interface that contains normalized paper metadata, linked repositories, and lightweight readiness signals for reproducibility. This interface serves as the input for subsequent modules, which attach experimental resources and proceed to initialization, evaluation, and optimization.

\subsubsection{External Resource Acquisition: Bridging the Dependency Gap}

While \textbf{AgentResource} successfully grounds papers to functional code repositories, the empirical utility of these artifacts is frequently bottlenecked by the absence of heavyweight external dependencies. Modern AI research, particularly as represented in recent conference proceedings, relies heavily on large-scale datasets, pre-trained weights, and specialized checkpoints that are typically referenced implicitly rather than hosted within the Git repositories themselves. To bridge this gap between code usability and execution readiness, we introduce the \textbf{External Resource Acquisition} module. This stage employs a decoupled, two-phase architecture---comprising \textit{Symbolic Resource Discovery} and \textit{Physical Download Execution}---to ensure resource-intensive tasks are handled with high efficiency and minimal failure rates.

\subsubsubsection{Task I: Symbolic Resource Discovery}
The initial phase focuses on the symbolic identification of dependencies, intentionally isolating metadata extraction from network-intensive execution. Operating under a strict \textbf{zero-download constraint}, the system parses the repository's internal structure and evaluation manifests (e.g., \texttt{rubric.csv}) to catalog external requirements. 

This process culminates in a \textbf{Global Resource Registry}, a centralized metadata ledger where each paper (identified by a unique sequence ID, $seq$) is mapped to its requisite assets. For each entry, the registry records critical attributes including the source URL, resource taxonomy (e.g., \texttt{dataset} vs. \texttt{model}), and estimated file size. We define the total projected workload as $S_{\text{total}} = \text{cal\_total\_size\_bytes}$, a metric that serves as a primary heuristic for downstream scheduling and computational resource allocation.

\subsubsubsection{Task II: Physical Acquisition and Orchestration}
In the second phase, the system transforms the symbolic registry into local physical assets. To ensure atomicity and reproducible state transitions, the \textbf{Download Execution Agent} operates exclusively on the validated registry without re-scanning source repositories. The execution logic is governed by four core mechanisms designed for industrial-grade robustness:

\begin{itemize}
    \item \textbf{Gated Selection and Filtering:} To optimize bandwidth and storage utilization, we implement a size-based gating mechanism. Only tasks where $S_{\text{total}}$ falls within a pre-defined range $[S_{\min}, S_{\max}]$ are admitted to the execution queue. Tasks with indeterminate or negative size estimates are preemptively discarded to prevent ``runaway'' downloads. Furthermore, a redundancy elimination layer cross-references historical logs to skip $seq$ IDs that have already reached a terminal state (success or failure).
    
    \item \textbf{Semantic Storage Architecture:} Resources are programmatically routed into a standardized, hierarchical directory structure. Assets are partitioned based on their identified type into \texttt{datasets/}, \texttt{models/}, \texttt{checkpoints/}, or \texttt{misc/}. To ensure cross-platform file system compatibility, filenames are sanitized by replacing illegal characters (e.g., \texttt{/}, \texttt{\textbackslash s}) and truncating paths to a maximum of 80 characters.
    
    \item \textbf{Polymorphic Retrieval Modalities:} The agent supports multiple retrieval strategies tailored to source complexity. For standard distribution channels, a \textbf{rule-based mode} utilizes pattern matching: recognized HuggingFace repositories are fetched via an optimized \texttt{huggingface-cli}, while generic HTTP/HTTPS links are managed via \texttt{curl} with stringent timeout bounds to prevent thread starvation. For non-standard or obfuscated sources, the system invokes an \textbf{LLM-driven autonomous mode} to generate and execute custom Python scripts capable of navigating complex download interfaces.
    
    \item \textbf{Concurrency and Fault Tolerance:} To maximize network throughput, we employ paper-level parallelism, enabling multiple tasks to progress concurrently while maintaining serial integrity within the downloads of an individual paper. An \textbf{asynchronous telemetry service} monitors real-time disk I/O and provides dynamic progress mapping. Reliability is further bolstered by a globally managed, in-memory URL mapping table that natively supports de-duplication and breakpoint resumption by filtering out previously successfully retrieved URLs from the active queue.
\end{itemize}

By the conclusion of this stage, the system has successfully transformed high-level paper citations into \textbf{Execution-Ready Task Units}. These units provide the synchronized combination of normalized code, datasets, and model weights required for the subsequent optimization and evaluation modules.

\subsection{AgentObjective: Automated Objective Rubric Construction}

In the current landscape of AI research replication, the formulation of high-quality evaluation rubrics is central to assessing the quality of replication quality. However, most existing benchmarks~\cite{starace2025paperbench, xiang2025scireplicate, yan-etal-2025-lmr} remain heavily dependent on manual labeling by domain experts or the direct involvement of original authors to define task objectives. While this expert-intensive paradigm is feasible for small-scale datasets, it creates a severe scalability bottleneck within the AutoSOTA framework, which requires large-scale, closed-loop optimization across vast quantities of academic literature. Furthermore, to overcome the \textit{diagnostic opacity} caused by ambiguous execution paths in long-horizon replication, the system must be capable of capturing complex intermediate logic and providing \textit{Dense Feedback}. This demand for extreme evaluative granularity further amplifies the inherent efficiency and cost defects of manual rubric construction, rendering it inadequate for the demands of massive-scale scientific discovery.

\begin{figure}
    \centering
    \includegraphics[width=\linewidth]{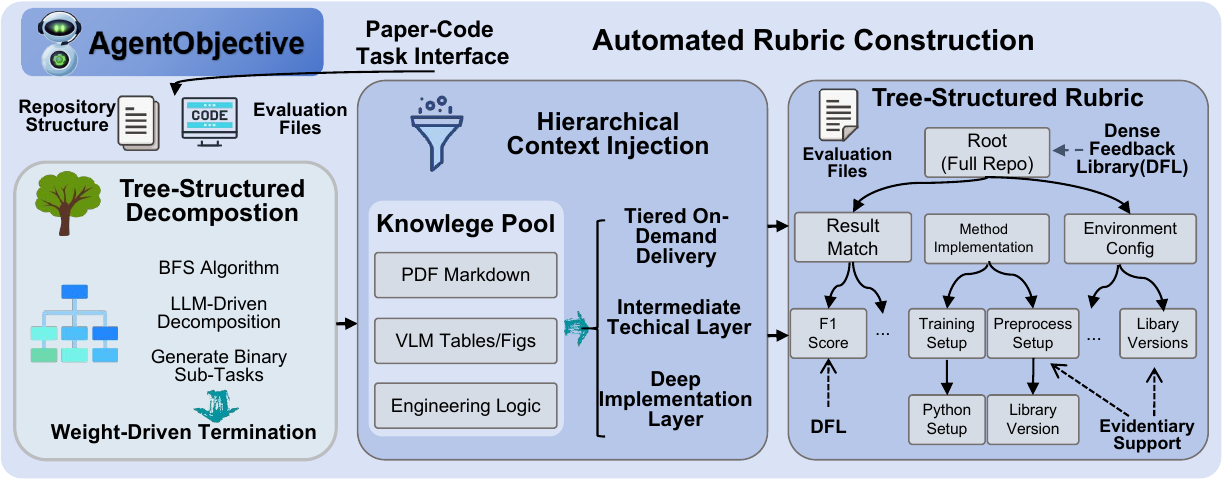}
    \caption{Overall Framework of the Rubric Construction Process}
    \label{fig:Rubric_framework}
\end{figure}

Therefore, we define \textbf{AgentObjective} as the pipeline stage responsible for the fully automated construction of a structured evaluation space prior to large-scale experiment replication and optimization. It no longer views a rubric as a simple, human-predefined scorecard, but rather employs a protocol-respecting approach to automatically map the macro research goal $g^{*}$ into a massive, quantifiable dense feedback library. Essentially, AgentObjective addresses the automated scaling challenge of transitioning from unstructured scientific literature $\mathcal{P}$ to a multi-level evaluation hierarchy. Without any human intervention, it precisely specifies the key results, technical nodes, and the logical constraints that must be observed between them, providing explicit evidentiary support for each atomic sub-task derived from both $\mathcal{P}$ and the ground-truth codebase $\mathcal{C}$.

\subsubsection{Tree-Structured Recursive Decomposition}

To render the complex and large-scale undertaking of research evaluation tractable, \textbf{AgentObjective} introduces a tree-structured rubric generation framework. By recursively decomposing macro-replication objectives into multi-dimensional, granular atomic sub-tasks, the agent achieves comprehensive coverage across the entire research lifecycle. In this structure, each parent node represents a high-level research module, such as \textit{Result Match}, \textit{Methodology Implementation}, \textit{Environment Configuration}, etc., while its children refine this logic into specific, verifiable facets. Consequently, when a replication failure occurs, the system can precisely localize the point of failure along the tree path, distinguishing where the issue arises from, providing critical diagnostic feedback for the iterative optimization and replication process of the Agent system.

The construction logic utilizes an automated algorithm based on Breadth-First Search (BFS) to systematically expand the tree-structured rubric. Starting from the root node, which represents the complete replication of the paper, \textbf{AgentObjective} leverages the reasoning capabilities of Large Language Models (LLMs) to dynamically decompose tasks into several binary (Pass/Fail) sub-tasks. To ensure evaluative depth while regulating the search space and enhancing computational efficiency, we implement a weight-driven termination mechanism. The root is assigned an initial total weight, which is strictly conserved and distributed during the downward decomposition process. Growth terminates once a node's assigned weight falls below a predefined threshold or the tree reaches its maximum logical depth. This mechanism ensures that the generated rubric remains focused on the paper’s core scientific contributions while maintaining a controllable and efficient granularity across papers of varying budgets and complexities.

\subsubsection{Hierarchical Context Injection}

To provide the constructed rubric with precise and sufficient evidentiary support while mitigating the risks of context overflow and noise interference inherent in long-context processing, \textbf{AgentObjective} implements a multi-dimensional knowledge injection pool. This pool serves as a centralized resource that integrates PDF Markdown data from the paper $\mathcal{P}$, visual knowledge extracted from tables and figures and the underlying engineering logic of the official codebase $\mathcal{C}$. By synthesizing these heterogeneous data sources, the agent eliminates the \textit{information chasm} that frequently exists between high-level academic descriptions and their low-level physical implementations, ensuring that every verification point in the rubric is anchored by a verifiable chain of evidence.

The core of this stage is a tiered \textit{on-demand} context delivery mechanism, dynamically triggered by the depth of the BFS decomposition to balance evaluative precision, reasoning stability, and token economic efficiency. This mechanism operates across three distinct logical layers:

\begin{itemize}

    \item \textbf{Shallow Strategic Layers}: At the apex of the tree-structured rubric, the system injects only macro-level information, such as the table of contents, abstract, and introduction. This prevents the model from being overwhelmed by premature technical details and guides it toward establishing a holistic strategic perception of the research objectives and core contributions.
    
    \item \textbf{Intermediate Technical Layers}: As decomposition progresses into algorithmic modules, detailed methodology and experimental sections are introduced alongside VLM-derived visual evidence. This layer strengthens the rigorous verification of experimental consistency and the alignment of target performance metrics $g^{*}$.
    
    \item \textbf{Deep Implementation Layers}: Upon reaching nodes concerning atomic engineering operations, the agent activates repository-level context. This recovers fine-grained implementation logic such as specific data pre-processing steps and utility configurations that may be omitted in the academic manuscript but remain essential for synthesizing a functional $\mathcal{R}_{rep}$.
    
\end{itemize}

Ultimately, this hierarchical injection framework of \textbf{AgentObjective} transforms the static knowledge within $\mathcal{P}$ and $\mathcal{C}$ into a dynamic, task-relevant stream of validated support, which ensures that the resulting rubric is not merely a structural template, but a high-fidelity diagnostic instrument grounded in the empirical reality of the research, thereby establishing a rigorous foundation for the subsequent replication and optimization of $\mathcal{R}_{rep}$.

In our current realization of AutoSOTA pipeline, the evaluation and refinement loops have primarily adopted a results-centric strategy, focusing heavily on \textit{Result Match} categories within the rubric, strongly emphasizing on the values of the proposed method and the best baseline method within paper $\mathcal{P}$. While effective for verifying that $\text{eval}(\mathcal{R}_{rep}) \approx g^{*}$, this approach may overlook the structural and semantic fidelity. Future work will expand the utility of \textbf{AgentObjective} by incorporating the remaining dimensions of the tree-structured rubric into the replication and optimization cycle. By utilizing these fine-grained categories as dense reward signals, we aim to step toward more valid replication and more profound structural innovations, providing more robust feedback necessary to bridge the chasm between theoretical innovation and empirical excellence in autonomous AI research.

\begin{figure}
    \centering
    \includegraphics[width=\linewidth]{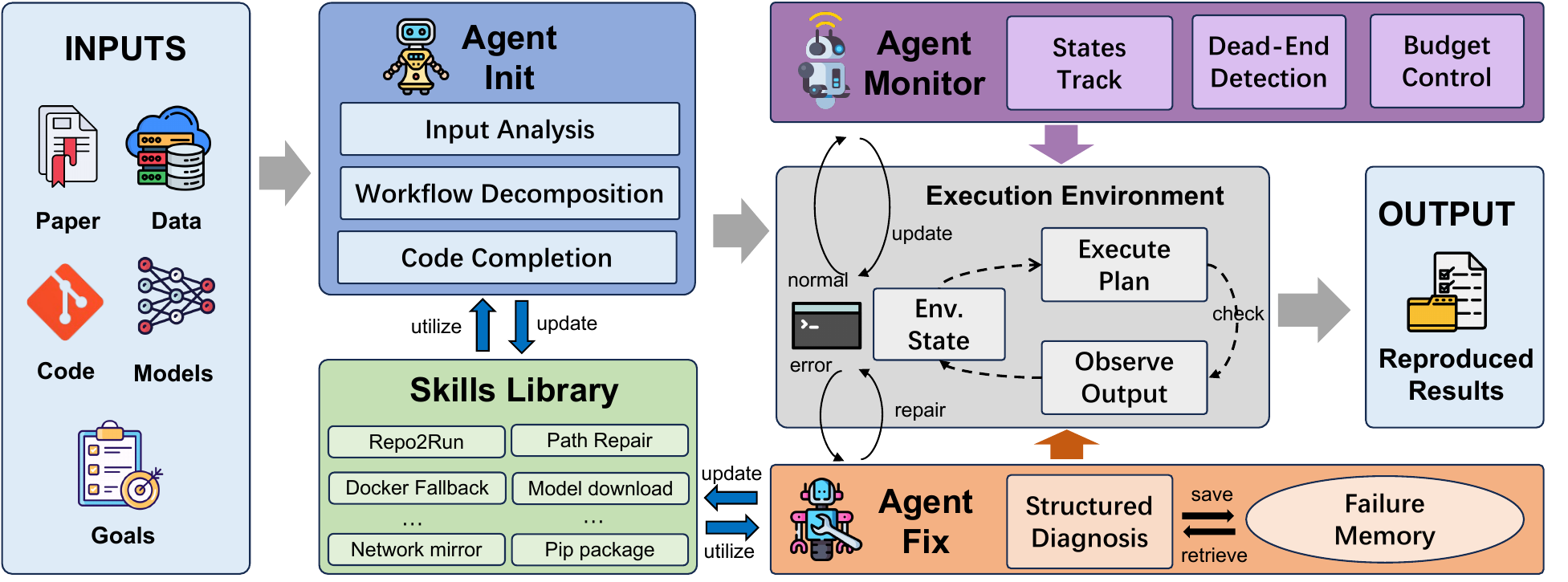}
    \caption{Overall Framework of the Replication Process}
    \label{fig:replication_framework}
\end{figure}

\subsection{AgentInit: Initialize Experiment Environment}

After \textbf{AgentResource} grounds papers to repositories and \textbf{AgentObjective} constructs the target rubric, the next challenge is to transform a noisy academic artifact into an executable experimental state. In practice, research repositories are rarely self-contained software products. They often contain undocumented environment assumptions, missing scripts, obsolete dependencies, incomplete training pipelines, hard-coded local paths, or evaluation commands that are only implicitly specified across the README, the paper PDF, and scattered code comments. As a result, initializing a reproducible run cannot be treated as a single environment setup step; it must be framed as a structured agentic process that converts a paper-grounded task interface into an actionable execution plan and a runnable repository state, see fig~\ref{fig:replication_framework}.

We therefore define \textbf{AgentInit} as the module responsible for workflow-grounded execution initialization. Given the grounded paper-task tuple produced upstream, AgentInit maps the paper, repository, resources, and rubric into an initial executable state:

\[
f_{\text{init}} : (\mathcal{P}, \mathcal{C}, \mathcal{D}, \mathcal{M}, g^{*}) \rightarrow (\mathcal{R}_{0}, \mathcal{E}_{0}, \Pi_{0}),
\]

where \(\mathcal{R}_{0}\) denotes the initialized repository state, \(\mathcal{E}_{0}\) denotes the validated execution environment, and \(\Pi_{0}\) denotes an explicit task plan for subsequent execution and debugging. Rather than allowing the agent to improvise freely from raw input, AgentInit constrains the initialization process through a preset workflow that decomposes the task into several ordered stages, including input analysis, sub-task decomposition, environment construction, dependency resolution, resource acquisition, repository inspection, command discovery, and execution preparation.

The key idea is to externalize initialization knowledge into a \emph{workflow-and-skill} interface. Concretely, the agent is first instructed to analyze the paper title, repository URL, local paper files, and rubric-defined target metrics, so that execution is anchored to the actual scientific objective rather than to superficial repository heuristics. It then decomposes initialization into explicit sub-tasks such as: identifying the correct repository commit, constructing the Docker environment, validating GPU and CUDA availability, checking Python and PyTorch compatibility, locating the main training and evaluation entrypoints, determining required datasets and pretrained checkpoints, and extracting the exact experimental configuration that corresponds to the reported baseline. This decomposition is important because academic repositories rarely fail in a single obvious place; instead, they exhibit cascades of small hidden assumptions that must be resolved in the correct order.

To make this process robust at scale, AgentInit is equipped with a reusable skill library. In the current system, these skills include at least the following capabilities: environment setup with Repo2Run, manual Docker fallback when automated build fails, container-level validation of disk space, GPU visibility, CUDA availability, and Python version compatibility, dependency installation with mirror-aware package resolution, dataset and base-model acquisition, repository inspection through README and file-tree analysis, code editing for path repair or glue logic synthesis, and execution-command recovery when the official instructions are incomplete. Importantly, these skills are not merely convenience tools; they serve as structured inductive biases that reduce the search space of initialization. Instead of treating every setup failure as a fresh reasoning problem, the agent is encouraged to invoke a known procedural template, which substantially improves stability on long-horizon execution tasks.

A further responsibility of AgentInit is to bridge the gap between the conceptual description in the paper and the incomplete engineering realization in the repository. In many cases, the repository does not fully expose the code path needed to reproduce the main result. Required utility files may be absent, preprocessing steps may be only implied, and evaluation scripts may assume local resources that do not exist in the current environment. AgentInit addresses this by permitting \emph{protocol-preserving repository repair}: the agent may synthesize missing glue logic, repair file-system assumptions, or reconstruct non-core scripts, but it must do so under the constraint that the original evaluation protocol, dataset split, and reported target setting remain unchanged. In this sense, AgentInit is not simply a deployment module; it is the stage that transforms fragile academic software into a minimally runnable but scientifically faithful experimental baseline.

Therefore, the output of AgentInit is not only a configured environment, but a fully instantiated execution interface for downstream replication. This interface includes a validated containerized runtime, a prepared repository state, an explicit execution plan, and a set of discovered commands and resources that define how the baseline experiment should be launched. By converting brittle and heterogeneous paper repositories into structured executable states, AgentInit provides the necessary substrate for reliable monitoring, debugging, and later-stage optimization.

\subsection{AgentMonitor: Tracking the state of experiment and preventing deadlock}

Even after a runnable initial state is constructed, autonomous paper replication remains an extreme long-horizon reasoning-and-acting problem. The agent must execute shell commands, inspect outputs, edit code, install dependencies, and iterate across many partially observed failure states. In such settings, a purely self-driven execution agent is prone to two characteristic pathologies: first, it can become trapped in local debugging loops, repeatedly attempting low-value fixes without making genuine progress; second, it may lose awareness of the global objective and over-focus on transient engineering details, such as endlessly repairing environment errors while never advancing toward the target evaluation. These failure modes are particularly acute in academic repositories, where setup and execution errors are noisy, ambiguous, and often causally entangled.

We therefore introduce \textbf{AgentMonitor}, an external supervisory agent that performs real-time state tracking and high-level intervention throughout the replication process. Formally, AgentMonitor operates over the execution trace of the main agent and produces supervisory actions:
\[
f_{\text{mon}} : \mathcal{T}_{0:t} \rightarrow (s_t, a_t^{\text{sup}}),
\]
where \(\mathcal{T}_{0:t}\) denotes the accumulated execution trace up to time \(t\), \(s_t\) denotes the inferred execution state, and \(a_t^{\text{sup}}\) denotes a supervisory action such as continue, resume with guidance, fallback, terminate, or rollback. The key design principle is that AgentMonitor does not replace the execution agent; instead, it remains \emph{outside} the main execution loop, observing the stream of actions and outputs and intervening only when signs of stagnation, ambiguity, or dead-end behavior emerge.

To fulfill this role, AgentMonitor integrates two complementary capabilities: (1) \emph{online execution trace interpretation}, which enables real-time detection of failure patterns and dead-end trajectories, and (2) \emph{persistent external state tracking}, which maintains structured, cross-iteration context beyond the limits of a single model invocation. The former ensures timely intervention during execution, while the latter provides the global memory necessary for consistent long-horizon reasoning.

\subsubsection{Execution Trace Monitoring and Deadlock Intervention}

In our current realization, AgentMonitor is tightly coupled to a structured execution trace. The execution agent runs in a streamed mode that exposes assistant outputs, tool invocations, execution results, and termination signals in real time. AgentMonitor continuously parses this trace to estimate the current phase of execution, such as environment setup, dependency installation, experiment launch, evaluation, reporting, or failure handling. This phase-aware tracking allows the monitor to distinguish productive exploration from pathological repetition. For example, repeated environment edits with no successful state transition, repeated installation attempts on the same failing dependency, or long stretches of output without progress can all be recognized as indicators of a dead-end trajectory.

Once such a state is detected, AgentMonitor generates \emph{high-level corrective guidance} rather than low-level patch instructions. This distinction is important. The monitor should steer the agent back toward a more promising region of the search space without collapsing the autonomy of the main agent into brittle hand-coded control. Typical supervisory actions include instructing the agent to switch from automated environment construction to manual Docker setup, to consult a dedicated failure-handling skill before making another installation attempt, to stop pursuing a low-probability fix and report failure for the current branch, or to resume execution from a clarified objective. This high-level intervention design preserves agent flexibility while still preventing unbounded local search.

AgentMonitor also serves as the system's \emph{global budget and safety controller}. Replication is subject to strict time and compute limits, and the system must remain responsive to stalled or runaway processes. Accordingly, the monitor tracks remaining wall-clock budget, limits the number of interaction rounds, terminates execution when timeout thresholds are crossed, and forces process-group cleanup when the run becomes unresponsive. In more advanced optimization settings, the same supervisory mechanism can trigger rollback to previously validated states when a new branch causes severe regression or persistent instability.

\subsubsection{Persistent Context Management via External Memory}

A fundamental challenge in long-horizon code optimization is managing state and context over extensive iterative processes. Each iteration involves a large volume of information, including the codebase, evaluation outputs, historical modification records, the current idea pool, and accumulated research insights, which collectively exceed the capacity of a single LLM context window.

To address this limitation, AgentMonitor maintains a \textbf{structured external memory architecture} that persists critical state in the form of local documents:

\begin{itemize}
    \item \textbf{\texttt{code\_analysis.md}}: A code cognition map constructed via a one-time comprehensive exploration of the target repository. It captures the full pipeline workflow, key source files and their responsibilities, evaluation procedures, script entry points, and immutable experimental constraints. This document serves as the primary entry point for subsequent code localization and reasoning.

    \item \textbf{\texttt{idea\_library.md}}: A dynamically evolving idea pool that records all candidate ideas, attempted modifications, and their outcomes. By externalizing the optimization trajectory, it enables consistent cross-iteration reasoning even when the model’s internal state is reset.

    \item \textbf{\texttt{research\_report.md}}: An external knowledge source populated prior to execution via a dedicated research phase. It provides distilled insights from relevant literature, including state-of-the-art techniques, empirically effective configurations, and concrete optimization strategies.
\end{itemize}

To further control context growth, the system employs two complementary mechanisms. First, a one-time distillation phase compresses the global structure of the repository into a persistent representation, eliminating the need for repeated full-repository scans. Second, during execution, the agent performs on-demand retrieval through command-line operations such as symbolic search, pattern matching, and targeted file inspection, allowing it to localize relevant code regions without loading the entire codebase into context.

Together, these mechanisms establish a retrieval-augmented paradigm for code optimization that tightly couples external memory with execution monitoring. This design not only bounds context usage within manageable limits, but also enables AgentMonitor to maintain a coherent global view of the optimization process across long horizons.

\subsection{AgentFix: Resolving Runtime Conflicts}

A robust replication system cannot rely on generic chain-of-thought reasoning alone to repair the diverse and recurring failures of academic software. In practice, many failure modes are not novel research problems but recurring engineering patterns: package installation timeouts, CUDA-toolchain mismatches, inaccessible model hubs, missing compiler toolchains, broken file paths, incompatible Python versions, absent checkpoints, corrupted caches, insufficient shared memory, or evaluation scripts that silently assume a different directory layout. If each failure is handled from scratch, the system wastes substantial budget rediscovering the same fixes across papers. This motivates a dedicated repair module that turns failure handling into an explicit, reusable component of the framework.

We define \textbf{AgentFix} as the module responsible for structured diagnosis, protocol-preserving repair, and cross-task reuse of failure knowledge. Given an execution state and an observed failure, AgentFix maps the failure into a validated repair action:

\[
f_{\text{fix}} : (\mathcal{R}_{t}, \mathcal{E}_{t}, \epsilon_t, \mathcal{M}_{\text{err}}) \rightarrow (\Delta_t, \mathcal{R}_{t+1}, \mathcal{M}_{\text{err}}^{\prime}),
\]

where \(\epsilon_t\) is the current failure signal, \(\Delta_t\) is the selected repair action, and \(\mathcal{M}_{\text{err}}\) is a reusable memory of historical failures and remedies. The emphasis here is not on unrestricted debugging, but on \emph{skill-augmented repair}: before attempting arbitrary modifications, the agent is required to retrieve a relevant failure-handling procedure from a structured skill repository whenever the error matches a known category.

In the current system, this design is already instantiated through explicit repair skills for high-frequency failure families. For example, package installation errors are handled through a dedicated \texttt{pip}-failure note that codifies known fixes such as using mirror-aware retries, installing version-compatible prebuilt wheels, downgrading incompatible build dependencies, cloning \texttt{git+https} packages outside the container, or switching to a development image when CUDA extensions require \texttt{nvcc}. Similarly, network and download failures are handled through a dedicated skill that encodes environment-specific constraints such as the lack of direct internet access inside Docker containers and the need to use mirror endpoints for HuggingFace, \texttt{pip}, and \texttt{conda}. This design principle can be summarized as \emph{retrieval before repair}: when the failure falls into a known operational class, the system first consults an explicit repair protocol rather than improvising from raw error messages.

Beyond these already implemented skills, the AgentFix abstraction naturally extends to a broader repair taxonomy, including GPU visibility and driver mismatches, Python-version migration, corrupted lockfiles, file-permission errors, disk exhaustion, shared-memory insufficiency, missing utility scripts, malformed path assumptions, broken symbolic links, evaluation-command mismatches, and partial repository reconstruction. Importantly, these repairs are not allowed to alter the scientific protocol itself. AgentFix may repair the engineering path to execution, but it may not change the evaluation semantics, the dataset split, or the methodological constraints that define comparability with the original paper. In this way, AgentFix complements AgentSupervisor: the former expands the space of admissible engineering repairs, while the latter bounds that space by scientific validity.

A second key component of AgentFix is its \textbf{failure memory mechanism}. Academic repositories exhibit strong error recurrence across papers, especially within the same ecosystem of libraries, CUDA versions, and download infrastructures. To exploit this regularity, AgentFix maintains reusable memory artifacts that record encountered failures, the repairs attempted, and the outcomes of those repairs. This memory can exist at multiple levels, including per-paper local memory for the current run, cross-paper global memory for recurring operational failures, and structured knowledge notes that distill repeated successes into explicit skills. Operationally, this enables the system to normalize raw tracebacks into reusable failure signatures, retrieve historically successful repair strategies, avoid repeating previously failed actions, and gradually convert one-off debugging episodes into stable procedural knowledge.

This memory-augmented repair loop is especially important for preventing cyclical reasoning. A common failure mode of autonomous agents is to oscillate among semantically equivalent fixes, such as repeatedly reinstalling an incompatible package, toggling between similar commands, or alternating between two broken paths without recognizing that both belong to the same unresolved root cause. By storing normalized error signatures and associating them with attempted remedies, AgentFix can explicitly rule out already-exhausted branches and steer the system toward unexplored repair strategies. Thus, memory does not merely improve efficiency; it changes the topology of the debugging process from blind local search into progressively informed diagnosis.

Therefore, AgentFix serves as the operational reliability layer of AutoSOTA. It turns failure handling from an ad hoc byproduct of agent execution into a first-class component with reusable skills, explicit repair protocols, and accumulating memory. In combination, these mechanisms significantly improve the agent's ability to recover from real-world software failures while preserving experimental comparability, thereby making large-scale autonomous replication substantially more stable and cost-effective.

\subsection{AgentIdeator: Constraint-Aware Hypothesis Construction}
\label{sec:agent_ideator}

Within our pipeline, ideation is the stage that constructs a structured search space of admissible improvement hypotheses before any expensive optimization loop is executed. Naive idea generation can easily devolve into unconstrained metric chasing, where invalid shortcuts are mixed with legitimate improvements. Such shortcuts may include implicit changes to evaluation logic, leakage from test data, output post-processing that violates the original protocol, or alterations that invalidate comparability with the paper baseline. 
Therefore, AgentIdeator is framed not as brainstorming, but as protocol-respecting structured hypothesis construction, providing a risk-aware hypothesis library along with an audit record that explains why each candidate is admissible, rejected, or requires human review. In essence, AgentIdeator specifies \emph{what may be optimized}, \emph{under which constraints}, and \emph{with what evidential support}.

The search beyond local code inspection is broadened by injecting external research expertise.
Instead of relying solely on the provided repository, this module queries an external, research-capable model to generate a task-relevant prior report grounded in SOTA literature patterns, empirical heuristics, and community best practices.
By synthesizing high-level insights from the broader field, this stage ensures that the candidate space includes profound structural innovations and domain-aware possibilities that are not immediately visible from the static codebase alone.

To ensure the generated ideas are both executable and scientifically valid, this layer anchors hypotheses to the concrete task instance through a multi-dimensional alignment process.
We enforce a red-line system to filter out any proposals that might violate the experimental integrity. Alignment is strictly maintained across four dimensions:
\textbf{Metric Alignment}: Hypotheses must target the specific objectives reported in the paper rather than generic performance metrics.
\textbf{Implementation Alignment}: Proposals must remain plausible relative to the actual capabilities of the reproduced environment and codebase.
\textbf{Constraint Alignment}: This preserves critical invariants, such as evaluation logic, dataset splits, and the core methodology's theoretical boundaries.
\textbf{Lever Alignment}: High-value optimization ``levers'' are identified to connect abstract knowledge to concrete intervention points in the code.

The final output of this stage is the systematized hypothesis library, a structured artifact that guides the downstream scheduler. Each hypothesis $h$ in the set $H$ is formally categorized by granularity $\tau(h) \in \{\text{PARAM, CODE, ALGO}\}$ and assigned a risk level $\rho(h) \in \{\text{LOW, MEDIUM, HIGH}\}$. We define an admissibility predicate $Adm(h)$ such that:
\begin{equation}
    Adm(h) = 
    \begin{cases} 
    1 & \text{if } h \text{ preserves evaluation integrity} \\
    0 & \text{if } h \text{ invokes prohibited shortcuts}
    \end{cases}
\end{equation}
The resulting admissible set $H_{adm} = \{h \in H : Adm(h) = 1\}$ ensures that the subsequent optimization phase focuses only on legitimate, high-potential research directions.
This systematic handoff reduces wasted experimentation and ensures that any achieved metric gains are both meaningful and reproducible.

\begin{figure}
    \centering
    \includegraphics[width=\linewidth]{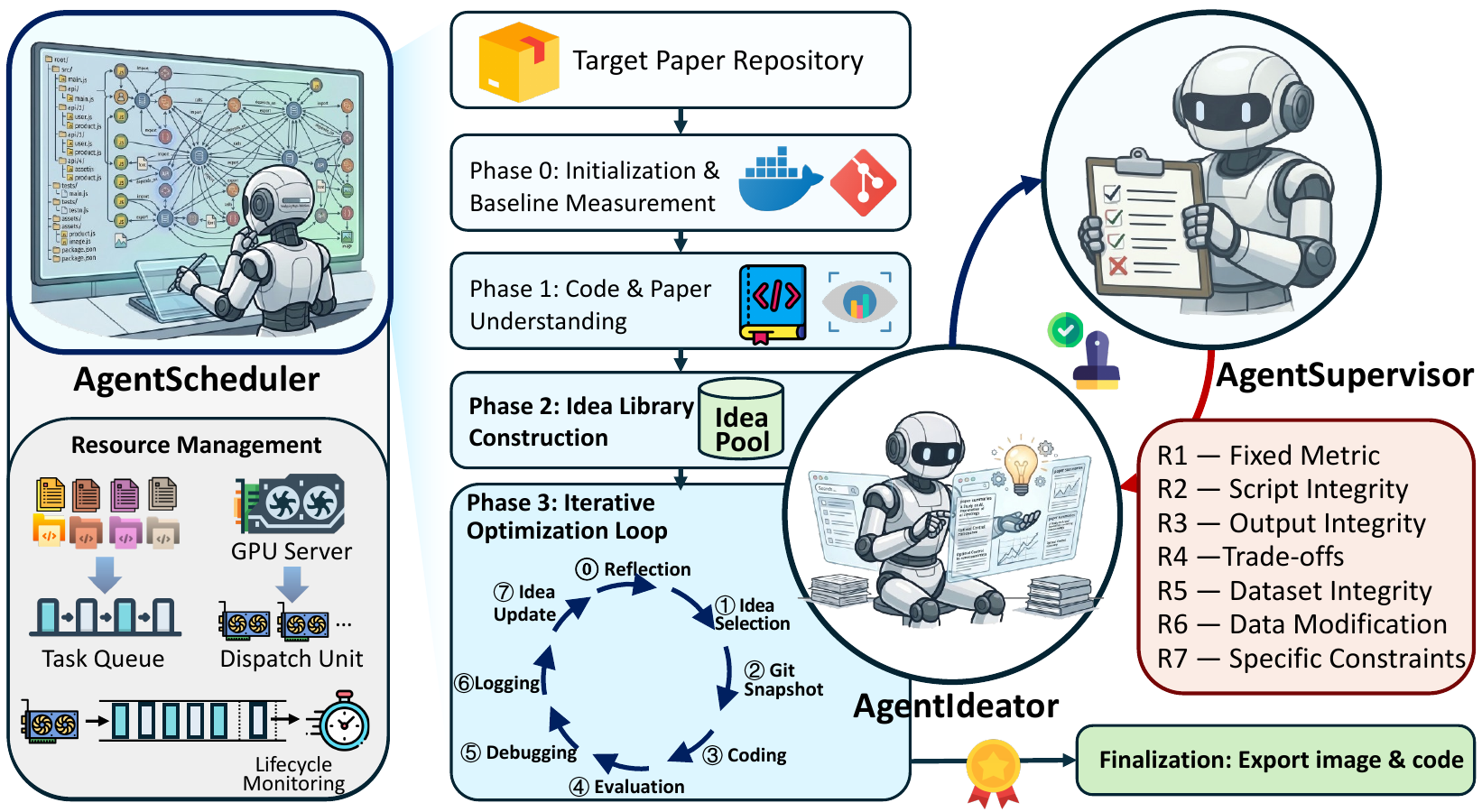}
    \caption{Overall Framework of the Reflection \& Ideation Process}
    \label{fig:optimize}
\end{figure}

\subsection{AgentScheduler: Lifecycle Management for Optimizing Research Codebases}

The AgentScheduler serves as the orchestration core of the optimization system, responsible for managing the complete lifecycle of a reproduced paper, from environment initialization and baseline measurement, to iterative code modification and evaluation, and finally to state persistence and result export. 
The Scheduler encompasses two key mechanisms: optimization strategy and resource management.

\subsubsection{Optimization Strategy}

The optimization workflow of the AgentScheduler is structured into four sequential phases: \textbf{Environment Initialization and Baseline Measurement} (Phase 0), \textbf{Code and Paper Understanding} (Phase 1), \textbf{Idea Library Construction} (Phase 2), and \textbf{Iterative Optimization Loop} (Phase 3). 
Compared with evolutionary optimization methods such as AlphaEvolve~\cite{novikov2025alphaevolve}, which target single algorithms or mathematical functions, the present system operates on a fundamentally more complex problem space. 
Unlike the well-defined and bounded search space of a single function in AlphaEvolve, where evaluation oracles are lightweight and deterministic, our system must optimize full research code repositories that are multi-file, multi-module, and highly interdependent. 
Each evaluation involves executing an entire machine learning experiment, subject to the strict boundaries of the original paper’s experimental setup. 
A more fundamental distinction lies in the candidate generation mechanism. 
While traditional evolutionary approaches can rely on exhaustive or massively parallel search, the optimization space of a research paper implementation is open and knowledge-intensive. 
Determining promising directions, assessing the potential impact of modifications, and introducing external techniques without violating core methodology all require expert-level reasoning.

The system leverages a LLM as a proposal generator, combining two essential capabilities: (i) deep understanding of the entire codebase, including cross-file dependencies, data flow across algorithmic stages, and evaluation protocol constraints; 
and (ii) broad awareness of state-of-the-art research, enabling the identification of structural bottlenecks and retrieval of relevant improvements from the literature. 
Both are indispensable: without code comprehension, proposals cannot be implemented; without research context, the search reduces to local parameter tuning with limited potential for substantial gains.

\textbf{Phase 0} executes essential initialization tasks. 
The optimizer launches a Docker container for the target paper, initializes a lightweight version control layer within the repository, and commits the unmodified code under the \texttt{\_baseline} tag. 
The full evaluation pipeline is then executed to obtain empirically measured baseline metrics. 
This step is critical, as reported results in the original paper often diverge from reproducible values due to software, hardware, or environmental differences. 
All subsequent optimization objectives are anchored to these measured baselines, which are recorded in a structured \texttt{scores.jsonl} ledger as iteration 0, providing a reproducible reference for later comparisons.

\textbf{Phase 1} systematically explores the repository structure, inspects evaluation scripts, identifies algorithmic branches, maps the full reasoning pipeline, and estimates runtime per evaluation. 
The output is a structured \texttt{code\_analysis.md}, serving as a cognitive map of the repository. 
This phase also requires explicit enumeration of \textit{hard constraints}, i.e., invariants that cannot be modified, which are recorded to guide subsequent agent decisions and to provide verifiable references for the AgentSupervisor.

\textbf{Phase 2} constructs the initial \textit{Idea Library}, a structured document listing candidate optimization directions. Each idea is annotated with type (\texttt{PARAM}/\texttt{CODE}/\texttt{ALGO}), priority, risk level, and implementation assumptions. 
Ideas are organized at three granularity levels: \textit{micro} (single parameter adjustments), \textit{meso} (function-level algorithmic improvements), and \textit{macro} (pipeline-stage restructuring). 
The initial library contains at least ten ideas, prioritized by feasibility and optimization potential, interleaving low-risk parameter tuning with higher-risk structural modifications. 
A \textit{redline audit} is conducted at the end of Phase 2, verifying each idea against the complete set of hard constraints and marking any violations as \texttt{REJECTED}. Only ideas labeled \texttt{CLEARED} proceed to execution scheduling.

\textbf{Phase 3} implements the iterative optimization loop, comprising seven mandatory steps per iteration: (0) Pre-Iteration Reflection, (1) Idea Selection, (2) Git Snapshot, (3) Code Implementation, (4) Evaluation, (5) Debugging, (6) Result Recording, and (7) Idea Library Update. 
The loop continues for up to \texttt{MAX\_ITERATIONS}, terminating early if the primary metric surpasses a predefined threshold.

A key design feature is the bifurcation between the \textit{Normal Path} and the \textit{Leap Path} during pre-iteration reflection. 
The agent inspects the types of ideas executed in the three most recent iterations: if all were \texttt{PARAM}-type adjustments, the current iteration is forced into the Leap Path, requiring a novel, structurally informed optimization idea derived from historical execution, rather than selection from existing ideas. 
This anti-stagnation mechanism prevents repeated exploration in low-dimensional parameter space and actively seeks structural breakthroughs when marginal gains from simple tuning are exhausted. 
Structural modifications introduced via the Leap Path are granted a \textit{Honeymoon Period}: if the initial result does not improve the best score, the system allows up to five subsequent iterations to explore and debug on the Leap-derived state before rolling back to the pre-Leap optimum. 
This ensures that potentially high-impact structural changes are adequately evaluated without premature rejection.

An internal \textit{version control layer} within the container underpins Scheduler correctness. 
Prior to any code modification (Step 2), the agent commits the current work tree and records the \texttt{PRE\_COMMIT} hash. In case of evaluation crashes or exhausted debug budget, a precise \texttt{git checkout} restores the previous state, preventing partial or contaminated changes from propagating. 
When a new optimum is found, the \texttt{record\_score.sh} script atomically advances the \texttt{\_best} tag to the corresponding commit. 
In the finalization phase (Phase 4), the container is restored to the \texttt{\_best} state, the final evaluation is executed, and the host generates a Docker image \texttt{autosota/paper:optimized} and exports the complete repository via \texttt{docker cp}, ensuring reproducibility and completeness of results.

\subsubsection{Resource Management}

At the system level, the AgentScheduler manages a global resource framework for multiple papers and multiple GPUs. 
The framework uses a two-GPU unit as the basic scheduling granularity, with each unit running at most one active task at any given time. 
Papers awaiting processing form a unified global task queue, and the scheduler continuously assigns tasks from the queue to available compute units in order.

Persistent tracking of the scheduling state is a key mechanism for ensuring reliable long-duration experiments. 
After each task assignment or completion, the AgentScheduler immediately writes the current state to disk, recording each paper's progress stage, assigned compute unit, process ID, and task start time. 
In the event of an unexpected server crash or forced termination of the scheduler, the system can restart in recovery mode, during which the scheduler automatically scans the persisted state, verifies whether each ``running'' process is still active, and infers the final results of terminated tasks by inspecting the corresponding output files. 
The paper status is updated accordingly, compute units are freed, and task dispatch resumes automatically, enabling uninterrupted continuation of multi-day or multi-week experiments without manual intervention.

Task execution is designed for long-term, unattended operation. Each task is launched in an independent process, fully decoupled from the scheduler, ensuring that ongoing tasks are unaffected if the user disconnects or the scheduler exits. 
For optimization tasks, the startup procedure automatically performs necessary preparation: if a paper has not completed its initial configuration, the system uses its replication Docker image and logs as context to trigger a configuration discovery process, extracting evaluation commands, baseline metrics, and environment parameters. 
GPU assignments are dynamically written into the paper's configuration to match the allocated compute unit, after which the optimization agent is launched autonomously. 
This design removes static bindings between compute units and papers, allowing any unit to immediately pick up the next queued task once the current one completes, achieving instant resource reuse without manual reconfiguration.

Upon task completion, the scheduler performs unified finalization: it verifies task success by checking the existence and content of output artifacts, automatically commits the optimized code as a new Docker image layer for successful tasks, and exports a host-accessible copy of the code. 
All containers created during the experiment are cleaned up, and GPU memory is released for subsequent tasks. 
This combination of dynamic assignment, artifact-driven completion verification, and immediate resource recycling ensures near-full utilization of compute resources throughout the experiment, enabling fully automated large-scale execution with zero manual intervention.

\subsection{AgentSupervisor: Preventing Invalid Optimization Behaviors}

The AgentSupervisor addresses a critical and non-trivial challenge in automated research systems: ensuring that long-horizon optimization proceeds under consistent, comparable, and methodologically valid conditions.
In our framework, the AgentScheduler continuously proposes and evaluates modifications to the codebase. 
However, without proper constraints, an unconstrained code-modifying agent may exploit unintended degrees of freedom and artificially inflate performance metrics, without improving the underlying method itself.
For instance, it may alter evaluation protocols, introduce test data leakage into training, relax dimensional or problem constraints, or even hard-code expected outputs. 
Such ``improvements'' are clearly unacceptable. In practice, our early experiments reveal that these behaviors occur more frequently than anticipated in autonomous optimization settings.
Therefore, the role of the AgentSupervisor is essential. 
It enforces comparability in a principled manner: any claimed performance gain must be evaluated under identical experimental settings, consistent methodological constraints, and the same evaluation pipeline.
Any optimization that violates these conditions is considered invalid and devoid of scientific value.


The core mechanism of the Agent Supervisor is the \textbf{Red Line System}, which is a set of six non-negotiable constraints that are enforced at multiple points in the optimization workflow:

\begin{itemize}
    \item \textbf{R1 — Evaluation Metric Parameters Must Not Change:} The parameterization of evaluation metrics must remain fixed. For retrieval tasks, the value of $k$ in recall@k must not be modified; for temporal or sequential tasks, the context window length and history window length used during evaluation must be preserved; aggregation strategies (such as averaging across multiple runs) must not be replaced by best-of-N or maximum value reporting.
    \item \textbf{R2 — Evaluation Script Integrity Must Be Maintained:} The evaluation script, score aggregation logic, and metric computation code must not be modified. Optimization operations must occur entirely upstream of the evaluation boundary.
    \item \textbf{R3 — Integrity of Algorithm Outputs Must Not Be Violated:} Model predictions and inference outputs must come from the actual model inference process. Overwriting, fabricating, or hard-coding predictions, even if it results in immediate metric gains, is strictly prohibited.
    \item \textbf{R4 — No Unfair Trade-offs Between Metric Dimensions:} The improvement in the primary optimization target metric must not come at the expense of significant degradation in other reported metrics. The system is required to report all metrics in each iteration, and cross-metric trade-offs are explicitly constrained.
    \item \textbf{R5 — Dataset Integrity Must Be Preserved:} The train/test split must strictly follow the definition in the original paper. Test data must not be incorporated into training in any form, including fine-tuning, calibration, or data augmentation.
    \item \textbf{R6 — No Modification of the Dataset:} Training or test data must not be altered, filtered, re-labeled, or re-sampled in any way that would change the evaluation distribution.
    \item \textbf{R7 — Paper-Specific Constraints Must Be Explicitly Identified and Strictly Followed:} Each paper's experimental setup introduces unique comparability requirements that may not be immediately evident from the general rules. For example, a temporal action prediction paper may have a specific history window length, which is part of the task definition itself; a text retrieval paper may define a context window size that influences the retrieval granularity. These key constraints must be explicitly identified and formalized as constraints before optimization begins.
\end{itemize}

These constraints are enforced not through single-point checks but through a multi-layered supervision mechanism that spans the entire optimization process.

\begin{itemize}
    \item \textbf{First Layer: Phase 1 Stage} — The agent is explicitly required to identify all relevant hard constraints for the specific paper at hand and enumerate them in a dedicated ``hard constraints/red line'' list in the \texttt{code\_analysis.md} document (i.e., generating R7). This forces the model to reason about the evaluation boundaries before generating any optimization ideas, rooting the constraint set in the specific context of the paper rather than relying solely on generic rules.
    \item \textbf{Second Layer: After Phase 2} — After the initial Idea library is generated, a mandatory \textbf{Red Line Audit} is performed. The agent must construct an audit table in the \texttt{idea\_library.md} document and verify each generated idea against all seven red lines. Any idea violating even a single constraint is immediately marked as \texttt{REJECTED (Red Line Violation)} and annotated with the violated rule. Only ideas marked as \texttt{CLEARED} can enter the Phase 3 execution cycle. This pre-execution filter ensures that no violating ideas are allowed to slip through into the optimization process.
    \item \textbf{Third Layer: When the Leap Path in Phase 3 is Triggered} — The idea synthesis process includes an explicit red line check for each candidate module. For each of the three innovative candidate solutions, the agent is required to check all eight constraints and discard any violating candidates. This secondary check is necessary because Leap ideas are dynamically generated and did not exist during Phase 2's audit.
    \item \textbf{Fourth Layer: Global Rule Section in Main Prompt} — The constraints are enforced as the highest-priority operational principles, stated as ``absolute prohibitions, with no exceptions under any circumstances.'' This phrasing is deliberate: it sends a signal to the model that these constraints take precedence over the primary optimization objectives, preventing the rationalization of violating constraints for the sake of pursuing better metrics.
\end{itemize}

The constraints of the AgentSupervisor serve as the boundaries for the search conducted by the AgentScheduler. This is not a conservative restriction on the agent's capabilities but a principled norm for defining ``what constitutes a legitimate improvement.'' Within these boundaries, the Agent Scheduler is free to explore; outside of these boundaries, regardless of the magnitude of metric improvements, the results are not considered valid.

\section{Experiments}

\subsection{Experiment Setup}
\label{subsec:experiment_setup}

To rigorously evaluate the capability of AutoSOTA in end-to-end autonomous replication and algorithmic optimization, we construct a highly curated evaluation benchmark sourced from recent top-tier AI conferences. We initialized our data collection by gathering accepted papers published within the past year across eight premier artificial intelligence conferences (including NeurIPS, ICLR, ICML, CVPR, ICCV, ACL, NAACL, and AAAI). 

To ensure the selected papers are suitable for automated, closed-loop optimization, we applied a systematic, multi-stage screening protocol. First, we conducted methodological filtering. We explicitly excluded purely theoretical studies, review articles, and purely analytical papers, isolating only those that propose a novel, empirically validated method, model, or algorithm. Second, we applied an artifact availability filter, retaining only papers that provided an explicit link to a public code repository. Third, to avoid wasting computational resources on incomplete releases, we performed a readiness assessment. This step strictly filtered out repositories that consisted solely of placeholders, empty directories, or ``code coming soon'' README announcements, ensuring that only repositories with actual implementation logic were retained. Following these initial resource acquisition stages, we assembled a preliminary pool of 2,347 candidate papers.

Due to the substantial computational overhead required for automated execution, we selected a subset of 745 papers from this pool for preliminary empirical testing. To maintain computational tractability and ensure that the iterative optimization loop could run efficiently at scale, we imposed strict environment and execution cost constraints. During this testing phase, we systematically filtered out papers that suffered from insurmountable environment installation failures or where a single complete experimental replication required more than 4 GPU hours to execute. 

Following this rigorous, multi-stage attrition process, we yielded a final, high-quality dataset of exactly 125 method-driven research papers. This diverse, multi-domain set of 125 papers serves as the primary empirical testbed for evaluating both the replication success rate and the subsequent SOTA-surpassing optimization capabilities of the AutoSOTA framework. Ultimately, as detailed in Section~\ref{subsec:main_result}, the system successfully optimized 105 of these 125 baselines, further demonstrating the robustness of our pipeline.

\subsection{Main Result}
\label{subsec:main_result}
Detailed experimental results are summarized in Table~\ref{tbl:main_4}. In this table, \textit{AutoSoTA Improvement} represents the performance gain achieved by AutoSOTA building upon the replicated main method of the original paper. Specifically, the reported improvement values imply the gains achieved on the single most scientifically significant core metric identified as the primary optimization target ($g^*$) for each respective paper.

Statistical analysis indicates that, provided all replicated results from the original main methods remain within reasonable confidence intervals, AutoSOTA not only achieves an exceptionally high automated replication success rate across diverse domains but also demonstrates superior potential for algorithmic optimization. In all tested cases, the improvement schemes automatically discovered by AutoSOTA yielded gains surpassing the original results reported in the papers, proving its capability to extract deeper performance augmentations through intelligent iteration even when operating on top-tier research foundations. Furthermore, from medical time-series analysis (e.g., ID 4) to protein binding energy prediction (e.g., ID 65), the system exhibits consistent robustness across multidisciplinary tasks, validating the generalizable potential of the framework to drive large-scale, automated, end-to-end scientific advancement while maintaining computational efficiency.

\definecolor{tablegray}{gray}{0.95}

\newcolumntype{L}[1]{>{\raggedright\arraybackslash}p{#1}}
\newcolumntype{C}[1]{>{\centering\arraybackslash}p{#1}}

\renewcommand{\arraystretch}{1.4}

\begin{longtable}{ C{0.8cm} L{6.5cm} C{2.5cm} C{2.5cm} }
    \caption{The Main Experiment of AutoSoTA} \label{tbl:main_4} \\
    
    \toprule
    \rowcolor[gray]{0.9} 
    \textbf{ID} & \textbf{Paper Title} & \textbf{Conference} & \textbf{Improvement} \\ \midrule
    \endfirsthead

    \multicolumn{4}{l}{{\textit{...Table \thetable{} continued from previous page}}} \\ \midrule
    \rowcolor[gray]{0.9}
    \textbf{ID} & \textbf{Paper Title} & \textbf{Conference} & \textbf{Improvement} \\ \midrule
    \endhead

    \midrule
    \multicolumn{4}{r}{{\textit{Continued on next page...}}} \\
    \endfoot

    \bottomrule
    \endlastfoot

    \showrowcolors 
    \rowcolors{1}{tablegray}{white}

    1 & SAVVY: Spatial Awareness via Audio-Visual LLMs through Seeing and Hearing~\cite{chen2026savvy} & NeurIPS2025 & 5.90\% \\
    2 & Pinet: Optimizing hard-constrained neural networks with orthogonal projection layers~\cite{grontas2025pinet} & ICLR2026 & 16.72\% \\
    3 & Discount Model Search for Quality Diversity Optimization in High-Dimensional Measure Spaces~\cite{tjanaka2026discount} & ICLR2026 & 7.32\% \\
    4 & Decentralized Attention Fails Centralized Signals: Rethinking Transformers for Medical Time Series~\cite{yu2026decentralized} & ICLR2026 & 4.45\% \\
    5 & Temporal Sparse Autoencoders: Leveraging the Sequential Nature of Language for Interpretability~\cite{bhalla2025temporal} & ICLR2026 & 2.25\% \\
    6 & PhySense: Sensor Placement Optimization for Accurate Physics Sensing~\cite{ma2026physense} & NeurIPS2025 & 4.16\% \\
    7 & Reasoning as Representation: Rethinking Visual Reinforcement Learning in Image Quality Assessment~\cite{zhao2025reasoning} & ICLR2026 & 2.68\% \\
    8 & Mean Flows for One-step Generative Modeling~\cite{geng2026mean} & NeurIPS2025 & 0.14\% \\
    9 & Score Matching with Missing Data~\cite{givens2025score} & ICML2025 & 0.14\% \\
    10 & Suitability Filter: A Statistical Framework for Classifier Evaluation in Real-World Deployment Settings~\cite{pouget2025suitability} & ICML2025 & 1.56\% \\
    11 & Orthogonal Subspace Decomposition for Generalizable AI-Generated Image Detection~\cite{yan2025orthogonal} & ICML2025 & 0.25\% \\
    12 & EfficientQAT: Efficient Quantization-Aware Training for Large Language Models~\cite{chen2025efficientqat} & ACL2025 & 6.08\% \\
    13 & APPL: A Prompt Programming Language for Harmonious Integration of Programs and Large Language Model Prompts~\cite{dong2025appl} & ACL2025 & 14.29\% \\
    14 & PIGuard: Prompt Injection Guardrail via Mitigating Overdefense for Free~\cite{li2025piguard} & ACL2025 & 3.66\% \\
    15 & FR-Spec: Accelerating Large-Vocabulary Language Models via Frequency-Ranked Speculative Sampling~\cite{zhao2025fr} & ACL2025 & 13.44\% \\
    16 & MathFusion: Enhancing Mathematical Problem-solving of LLM through Instruction Fusion~\cite{pei2025mathfusion} & ACL2025 & 20.97\% \\
    17 & Synergizing LLMs with Global Label Propagation for Multimodal Fake News Detection~\cite{hu2025synergizing} & ACL2025 & 0.83\% \\
    18 & CoT-based Synthesizer: Enhancing LLM Performance through Answer Synthesis~\cite{zhang2025cot} & ACL2025 & 3.68\% \\
    19 & Dynamic Scaling of Unit Tests for Code Reward Modeling \cite{ma2025dynamic}& ACL2025 & 2.12\% \\
    20 & Aristotle: Mastering Logical Reasoning with A Logic-Complete Decompose-Search-Resolve Framework\cite{xu2025aristotle} & ACL2025 & 1.68\% \\
    21 & Turning Trash into Treasure: Accelerating Inference of Large Language Models with Token Recycling\cite{luo2025turning} & ACL2025 & 19.50\% \\
    22 & Chain-of-Reasoning: Towards Unified Mathematical Reasoning in Large Language Models via a Multi-Paradigm Perspective\cite{yu2025chain} & ACL2025 & 3.38\% \\
    23 & Don’t Reinvent the Wheel: Efficient Instruction-Following Text Embedding based on Guided Space Transformation\cite{feng2025don} & ACL2025 & 17.50\% \\
    24 & Enhancing Automated Interpretability with Output-Centric Feature Descriptions \cite{gur2025enhancing}& ACL2025 & 2.69\% \\
    25 & DEEPER Insight into Your User: Directed Persona Refinement for Dynamic Persona Modeling\cite{chen2025deeper} & ACL2025 & 21.78\% \\
    26 & A Generative Adaptive Replay Continual Learning Model for Temporal Knowledge Graph Reasoning\cite{zhang2025generative} & ACL2025 & 12.40\% \\
    27 & CiteEval: Principle-Driven Citation Evaluation for Source Attribution\cite{xu2025citeeval} & ACL2025 & 24.10\% \\
    28 & Segment-Based Attention Masking for GPTs\cite{katz2025segment} & ACL2025 & 0.22\% \\
    29 & Conditional Dichotomy Quantification via Geometric Embedding\cite{cui2025conditional} & ACL2025 & 14\% \\
    30 & An Efficient and Precise Training Data Construction Framework for Process-supervised Reward Model in Mathematical Reasoning\cite{sun2025efficient} & ACL2025 & 1.12\% \\
    31 & Circuit Stability Characterizes Language Model Generalization \cite{sun2025circuit}& ACL2025 & 35.80\% \\
    32 & Personal Travel Solver: A Preference-Driven LLM-Solver System for Travel Planning \cite{shao2025personal}& ACL2025 & 4.48\% \\
    33 & Enhancing Unsupervised Sentence Embeddings via Knowledge-Driven Data Augmentation and Gaussian-Decayed Contrastive Learning\cite{lai2025enhancing} & ACL2025 & 2.81\% \\
    34 & Ensemble Watermarks for Large Language Models\cite{niess2025ensemble} & ACL2025 & 3.46\% \\
    35 & Comparing Moral Values in Western English-speaking societies and LLMs with Word Associations \cite{xiang2025comparing}& ACL2025 & 8.59\% \\
    36 & Synergistic Weak-Strong Collaboration by Aligning Preferences \cite{jiao2025synergistic}& ACL2025 & 28.42\% \\
    37 & Mitigating Confounding in Speech-Based Dementia Detection through Weight Masking~\cite{sheng-etal-2025-mitigating} & ACL2025 & 2.45\% \\
    38 & TinySAM: Pushing the Envelope for Efficient Segment Anything Model~\cite{10.1609/aaai.v39i19.34255} & AAAI2025 & 0.90\% \\
    39 & CALF: Aligning LLMs for Time Series Forecasting via Cross-modal Fine-Tuning~\cite{10.1609/aaai.v39i18.34082} & AAAI2025 & 0.19\% \\
    40 & Granite Guardian: Comprehensive LLM Safeguarding~\cite{padhi-etal-2025-granite} & NAACL2025 & 1.37\% \\
    41 & Auto-Regressive Moving Diffusion Models for Time Series Forecasting~\cite{10.1609/aaai.v39i16.33838} & AAAI2025 & 1.07\% \\
    42 & xPatch: Dual-Stream Time Series Forecasting with Exponential Seasonal-Trend Decomposition~\cite{10.1609/aaai.v39i19.34270} & AAAI2025 & 6.92\% \\
    43 & VHM: Versatile and Honest Vision Language Model for Remote Sensing Image Analysis~\cite{10.1609/aaai.v39i6.32683} & AAAI2025 & 0.67\% \\
    44 & Elevating Flow-Guided Video Inpainting with Reference Generation~\cite{10.1609/aaai.v39i3.32255} & AAAI2025 & 3.40\% \\
    45 & Unlocking the Power of LSTM for Long Term Time Series Forecasting~\cite{10.1609/aaai.v39i11.33303} & AAAI2025 & 0.86\% \\
    46 & Battling the Non-stationarity in Time Series Forecasting via Test-time Adaptation~\cite{10.1609/aaai.v39i17.33965} & AAAI2025 & 2.27\% \\
    47 & TimePFN: Effective Multivariate Time Series Forecasting with Synthetic Data~\cite{10.1609/aaai.v39i19.34288} & AAAI2025 & 8.77\% \\
    48 & Proxy-SPEX: Sample-Efficient Interpretability via Sparse Feature Interactions in LLMs~\cite{butler2026proxyspex} & NeurIPS2025 & 30.30\% \\
    49 & Hogwild! Inference: Parallel LLM Generation via Concurrent Attention~\cite{rodionov2026hogwild} & NeurIPS2025 & 4\% \\
    50 & CausalPFN: Amortized Causal Effect Estimation via In-Context Learning~\cite{balazadeh2026causalpfn} & NeurIPS2025 & 15.86\% \\
    51 & FlashTP: Fused, Sparsity-Aware Tensor Product for Machine Learning Interatomic Potentials~\cite{lee2025flashtp} & ICML2025 & 0.70\% \\
    52 & Non-stationary Diffusion For Probabilistic Time Series Forecasting~\cite{ye2025nonstationary} & ICML2025 & 1.28\% \\
    53 & $K^{2}$VAE: A Koopman-Kalman Enhanced Variational AutoEncoder for Probabilistic Time Series Forecasting~\cite{wu2025kvae} & ICML2025 & 1.52\% \\
    54 & TimeBase: The Power of Minimalism in Efficient Long-term Time Series Forecasting~\cite{huang2025timebase} & ICML2025 & 0.36\% \\
    55 & CSBrain: A Cross-scale Spatiotemporal Brain Foundation Model for EEG Decoding~\cite{NEURIPS2025_7e199ad8} & NeurIPS2025 & 6.25\% \\
    56 & InfoSAM: Fine-Tuning the Segment Anything Model from An Information-Theoretic Perspective~\cite{10.5555/3780338.378341infosam} & ICML2025 & 1.60\% \\
    57 & MDReID: Modality-Decoupled Learning for Any-to-Any Multi-Modal Object Re-Identification~\cite{feng2026mdreid} & NeurIPS2025 & 14\% \\
    58 & Mind-the-Glitch: Visual Correspondence for Detecting Inconsistencies in Subject-Driven Generation~\cite{eldesokeymind} & NeurIPS2025 & 1.80\% \\
    59 & Not All Data are Good Labels: On the Self-supervised Labeling for Time Series Forecasting~\cite{yangnot} & NeurIPS2025 & 0.30\% \\
    60 & IA-GGAD: Zero-shot Generalist Graph Anomaly Detection via Invariant and Affinity Learning~\cite{zhang2026ia} & NeurIPS2025 & 1.83\% \\
    61 & Hierarchical Shortest-Path Graph Kernel Network~\cite{wang2026hierarchical} & NeurIPS2025 & 2.20\% \\
    62 & VisionTS: Visual Masked Autoencoders Are Free-Lunch Zero-Shot Time Series Forecasters~\cite{10.5555/3780338.3780679VISION} & ICML2025 & 0.80\% \\
    63 & TS-RAG: Retrieval-Augmented Generation based Time Series Foundation Models are Stronger Zero-Shot Forecaster~\cite{ningtsRAG} & NeurIPS2025 & 13.90\% \\
    64 & Improving Time Series Forecasting via Instance-aware Post-hoc Revision~\cite{liuimproving} & NeurIPS2025 & 10.60\% \\
    65 & Predicting mutational effects on protein binding from folding energy~\cite{10.5555/3780338.3780841Predicting} & ICML2025 & 15.48\% \\
    66 & Tropical Attention: Neural Algorithmic Reasoning for Combinatorial Algorithms~\cite{hashemitropical} & NeurIPS2025 & 15.55\% \\
    67 & KAN-AD: Time Series Anomaly Detection with Kolmogorov-Arnold Networks~\cite{10.5555/3780338.3783524KANAD} & ICML2025 & 0.89\% \\
    68 & SEMPO: Lightweight Foundation Models for Time Series Forecasting~\cite{hesempo} & NeurIPS2025 & 0.12\% \\
    69 & Tree Ensemble Explainability through the Hoeffding Functional Decomposition and TreeHFD Algorithm~\cite{benardtree} & NeurIPS2025 & 36.60\% \\
    70 & Certified Unlearning for Neural Networks~\cite{10.5555/3780338.3781566Certified} & ICML2025 & 63.64\% \\
    71 & Neural MJD: Neural Non-Stationary Merton Jump Diffusion for Time Series Prediction~\cite{gaoneural} & NeurIPS2025 & 19\% \\
    72 & One Arrow, Two Hawks: Sharpness-aware Minimization for Federated Learning via Global Model Trajectory~\cite{DBLP:conf/icml/LiLCHL25} & ICML2025 & 4.37\% \\
    73 & Regularized Langevin Dynamics for Combinatorial Optimization~\cite{DBLP:conf/icml/Feng025} & ICML2025 & 0.24\% \\
    74 & Least squares variational inference~\cite{NEURIPS2025_d51ceada} & NeurIPS2025 & 0.02\% \\
    75 & Tree-Sliced Entropy Partial Transport~\cite{NEURIPS2025_181a0279} & NeurIPS2025 & 0.50\% \\
    76 & AANet: Virtual Screening under Structural Uncertainty via Alignment and Aggregation~\cite{NEURIPS2025_bf684916} & NeurIPS2025 & 6.79\% \\
    77 & Advancing Constrained Monotonic Neural Networks: Achieving Universal Approximation Beyond Bounded Activations~\cite{DBLP:conf/icml/SartorSS25} & ICML2025 & 0.30\% \\
    78 & Conformal Anomaly Detection in Event Sequences~\cite{DBLP:conf/icml/Zhang000LP25} & ICML2025 & 0.11\% \\
    79 & Differentially Private Federated $k$-Means Clustering with Server-Side Data~\cite{DBLP:conf/icml/ScottLS25} & ICML2025 & 2.01\% \\
    80 & Latent Score-Based Reweighting for Robust Classification~\cite{DBLP:conf/icml/TongZTGHL0K25} & ICML2025 & 6.58\% \\
    81 & Meta-Black-Box-Optimization through Offline Q-function Learning~\cite{DBLP:conf/icml/MaCJGG25} & ICML2025 & 1.01\% \\
    82 & Efficient Training-Free Online Routing for High-Volume Multi-LLM Serving~\cite{NEURIPS2025_c878e34a} & NeurIPS2025 & 1.61\% \\
    83 & STaRFormer: Semi-Supervised Task-Informed Representation Learning via Dynamic Attention-Based Regional Masking for Sequential Data~\cite{NEURIPS2025_1eb7e710} & NeurIPS2025 & 3.15\% \\
    84 & Towards Accurate Time Series Forecasting via Implicit Decoding~\cite{NEURIPS2025_0e82ef0c} & NeurIPS2025 & 2.30\% \\
    85 & NeuralSurv: Deep Survival Analysis with Bayesian Uncertainty Quantification~\cite{NEURIPS2025_4985e71a} & NeurIPS2025 & 23\% \\
    86 & On the Integration of Spatial-Temporal Knowledge: A Lightweight Approach to Atmospheric Time Series Forecasting~\cite{NEURIPS2025_7691484a} & NeurIPS2025 & 2.42\% \\
    87 & Multi-Task Vehicle Routing Solver via Mixture of Specialized Experts under State-Decomposable MDP~\cite{NEURIPS2025_2ee1c872} & NeurIPS2025 & 51.40\% \\
    88 & Channel Normalization for Time Series Channel Identification~\cite{DBLP:conf/icml/LeeP025} & ICML2025 & 15.20\% \\
    89 & Distinguishing Cause from Effect with Causal Velocity Models~\cite{xi2025distinguishing} & ICML2025 & 1.66\% \\
    90 & Information Bottleneck-guided MLPs for Robust Spatial-temporal Forecasting~\cite{chen2025information} & ICML2025 & 0.38\% \\
    91 & Learning Time-Aware Causal Representation for Model Generalization in Evolving Domains~\cite{he2025learning} & ICML2025 & 4.30\% \\
    92 & Modified K-means Algorithm with Local Optimality Guarantees~\cite{li2025modified} & ICML2025 & 0.84\% \\
    93 & FedWMSAM: Fast and Flat Federated Learning via Weighted Momentum and Sharpness-Aware Minimization~\cite{li2026fedwmsam} & NeurIPS2025 & 1.77\% \\
    94 & BounDr.E: Predicting Drug-likeness via Biomedical Knowledge Alignment and EM-like One-Class Boundary Optimization~\cite{bang2025boundre} & ICML2025 & 1.27\% \\
    95 & Accelerating Feature Conformal Prediction via Taylor Approximation~\cite{tang2026accelerating} & NeurIPS2025 & 5.49\% \\
    96 & Multi-Class Support Vector Machine with Differential Privacy~\cite{park2026multiclass} & NeurIPS2025 & 2.37\% \\
    97 & Wasserstein Transfer Learning~\cite{zhang2026wasserstein} & NeurIPS2025 & 3.74\% \\
    98 & X-Mahalanobis: Transformer Feature Mixing for Reliable OOD Detection~\cite{wei2026xmahalanobis} & NeurIPS2025 & 1.07\% \\
    99 & Fast Non-Log-Concave Sampling under Nonconvex Equality and Inequality Constraints with Landing~\cite{jeon2026fast} & NeurIPS2025 & 4.83\% \\
    100 & Iterative Missing Data Imputation with Model Form Adaptation and Non-Missing Feature Supervision~\cite{wang2026iterative} & NeurIPS2025 & 2.50\% \\
    101 & Measure-Theoretic Anti-Causal Representation Learning~\cite{behnam2026measuretheoretic} & NeurIPS2025 & 0.57\% \\
    102 & Stochastic Forward-Forward Learning through Representational Dimensionality Compression~\cite{zhu2026stochastic} & NeurIPS2025 & 1.33\% \\
    103 & Distributed Conformal Prediction via Message Passing~\cite{wen2025distributed} & ICML2025 & 6.33\% \\
    104 & Balanced Active Inference~\cite{chen2026balanced} & NeurIPS2025 & 24.80\% \\
    105 & Aligning Evaluation with Clinical Priorities: Calibration, Label Shift, and Error Costs~\cite{flores2026aligning} & NeurIPS2025 & 2.40\% \\
\end{longtable}

\begin{figure}
    \centering
    \includegraphics[width=\linewidth]{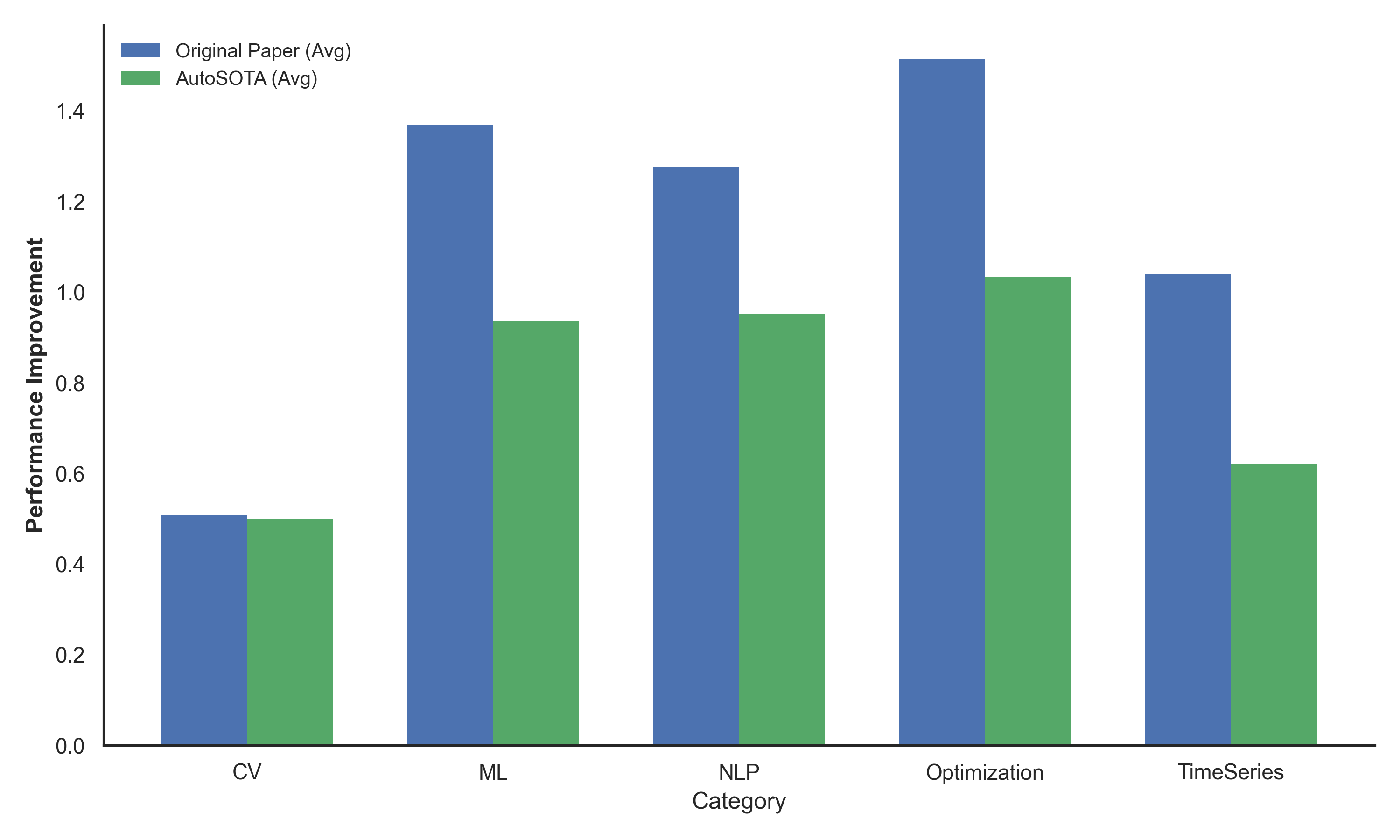}
    \caption{Comparison of Average Performance Improvements across Major Research Categories, illustrating the average performance enhancement of the original papers versus the AutoSOTA framework across five key domains: CV, ML, NLP, Optimization, and TimeSeries. To better visualize the relative gains across diverse scales of improvement, the vertical axis represents values processed via a $\log_{10}$ transformation ($y' = \log_{10}(y)$).}
    \label{fig:performance_comparison}
\end{figure}

Based on the themes of the conference papers utilized in our experiments, we selected five representative categories, including CV, ML, NLP, Optimization, and TimeSeries for in-depth analysis, as illustrated in Figure~\ref{fig:performance_comparison}. In terms of data processing, a $\log_{10}$ transformation was applied to the raw performance improvement data, which exhibits a significant order-of-magnitude span ranging from 3\% to 33\%. This approach enables smooth cross-domain comparisons and eliminates visual bias caused by extreme outliers. The analytical results indicate that AutoSOTA framework demonstrates remarkably robust optimization potential across all selected heterogeneous domains. Notably, in the CV domain, the performance gains achieved align closely with the average levels of the original papers, reflecting superior automation efficiency. Furthermore, in domains with higher gain potential, such as Optimization and ML, AutoSOTA maintains significant growth on the logarithmic scale, consistent with the trends observed in the original research. Overall, compared to traditional research data characterized by high volatility, AutoSOTA provides a performance benchmark with lower variance and stronger consistency. This fully validates the framework's scalability and generalization value when handling complex, interdisciplinary scientific research tasks.

\begin{figure*}[htbp]
    \centering
    \begin{subfigure}[b]{0.45\textwidth} 
        \centering
        \includegraphics[height=5cm, keepaspectratio]{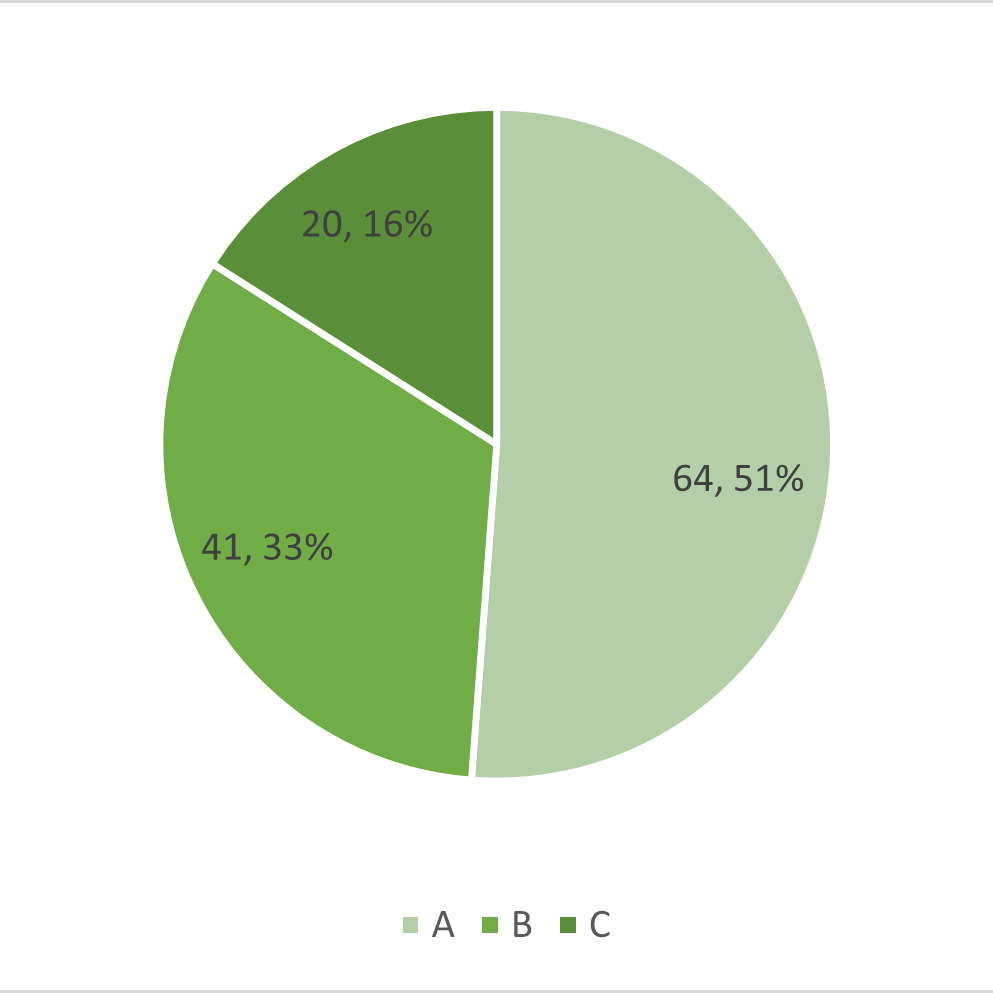} 
        \caption{Categorical distribution of AutoSOTA experiments. Class A (light green) represents deep optimizations involving algorithmic restructuring or architectural updates, accounting for 51\% (64/125) of total cases. Class B (medium green) denotes empirical hyperparameter tuning (41, 33\%), and Class C (dark green) indicates instances where no improvement was achieved.} 
        \label{fig:ratio_dist}
    \end{subfigure}
    \hfill 
    \begin{subfigure}[b]{0.5\textwidth}
        \centering
        \includegraphics[height=5cm, keepaspectratio]{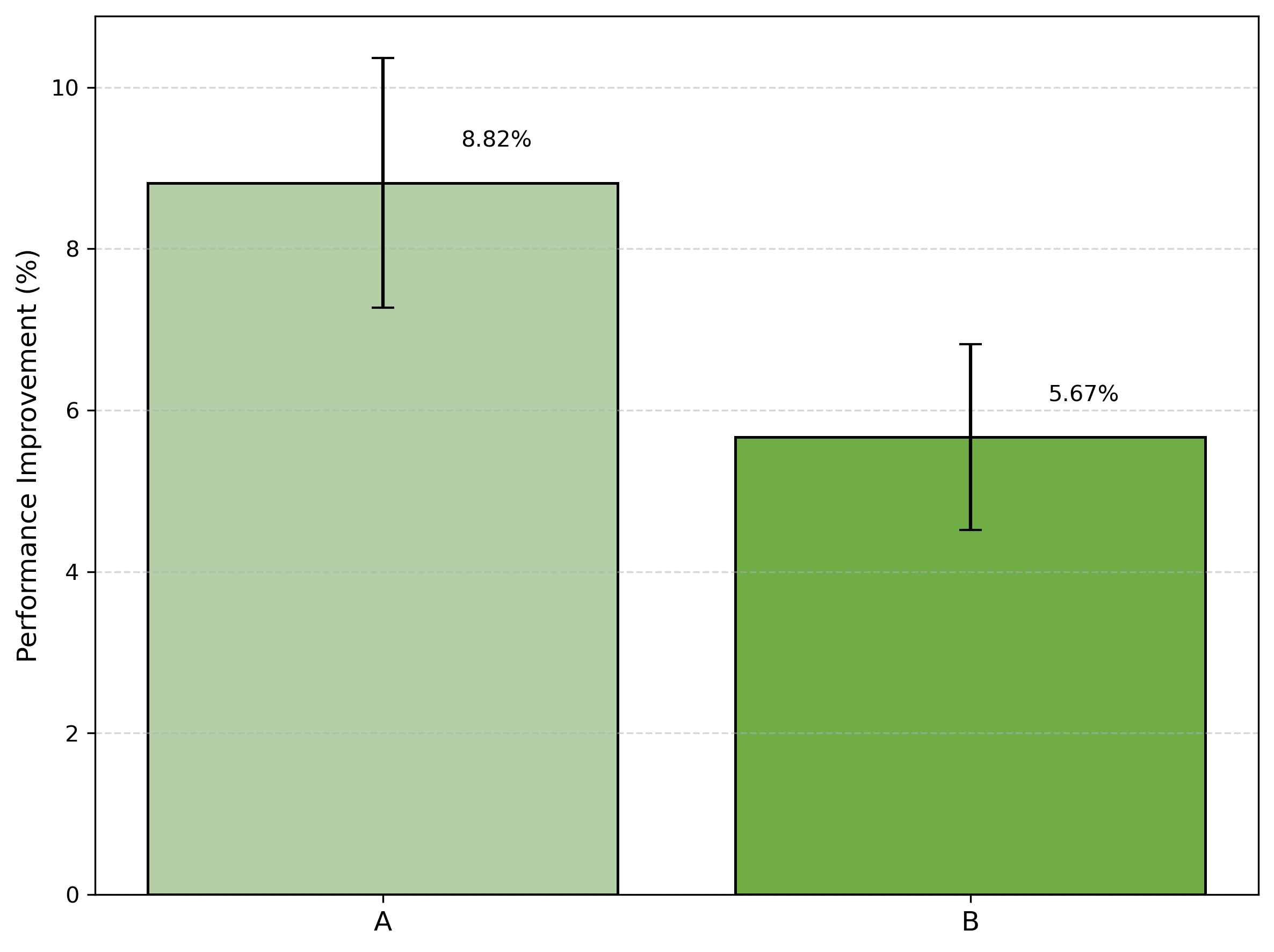} 
        \caption{Comparative performance between Class A (Algorithmic Innovation, light green) and Class B (Parameter Optimization, medium green). Class A algorithmic innovations yielded an average improvement of 8.82\%, significantly exceeding the 5.67\% mean improvement observed in Class B parameter tuning. Bar heights represent the mean, and error bars denote the Standard Error of the Mean (SEM).} 
        \label{fig:performance_percent}
    \end{subfigure}
    
    \vspace{1em}
    \caption{Duo-dimensional statistical analysis of AutoSOTA optimization outcomes} 
    \label{fig:combined_optimization_stats}
\end{figure*}

To elucidate the nature of the improvements generated by AutoSOTA, a further systematic audit and categorical analysis were conducted across 125 experimental trials ~\ref{fig:combined_optimization_stats}. Outcomes were strictly classified into three tiers: 
\textbf{Class A (Algorithmic Innovation)}, comprising substantive methodological modifications such as loss function redesign, model operator replacement, and even architectural restructuring; 
\textbf{Class B (Parameter Optimization)}, involving the fine-tuning of hyperparameters such as learning rates and sampling scales; 
and \textbf{Class C (Optimization Failure)}, where no better improvements were recorded over the original method.

The empirical evidence strongly validates the core claims of the AutoSOTA framework. 
First, regarding the distribution of improvements, \textbf{Class A} schemes involving substantive algorithmic innovation account for 51\% (64 out of 125) of the results. 
This confirms that the system is capable of logical restructuring rather than simply a conventional machine learning tool. 
Second, in terms of performance yield, \textbf{Class A} achieved a mean enhancement of 8.82\%, significantly outperforming the 5.67\% average observed in \textbf{Class B}. 
This statistical result demonstrates AutoSOTA's ability to identify and transcend fundamental algorithmic bottlenecks.

Notably, among the 125 paper baselines executed, 20 cases, taking $16\%$ out of all papers, suffer from invalid optimizations, denoted as \textbf{Class C}. These cases correspond to scenarios where the AgentSupervisor failed to detect or prevent potential violations during the optimization process. This result indicates that, in large-scale open environments, self-supervised constraint mechanisms still exhibit a certain failure rate, which may arise from implicit violations or limitations in handling complex cross-agent interactions. Despite these challenges, the overall results show that approximately $84\%$ of tasks maintain methodological consistency and evaluation comparability under the supervision mechanism. This suggests that the AgentSupervisor is effective in constraining the optimization process in most cases, ensuring that the 105 successfully optimized results retain a high level of experimental credibility.

\subsection{Case Study}
\subsubsection{Case Study for LLM Research.}

We use \textit{FR-Spec: Accelerating Large-Vocabulary Language Models via Frequency-Ranked Speculative Sampling} as the representative LLM case because it shows how AutoSOTA can find implementation-level optimization opportunities in a complex inference system. The main gain in this case does not come from routine hyperparameter tuning. Instead, AutoSOTA identifies a hidden error in the speculative decoding path and turns that finding into a clear throughput improvement. FR-Spec studies efficient speculative decoding for large-vocabulary language models. We evaluate it with MT-Bench throughput, measured in tokens per second. Starting from the released implementation, AutoSOTA establishes a baseline of 674.58 tok/s.

The optimization focuses on the speculative decoding loop. After reproducing the baseline, AutoSOTA detects abnormal acceptance behavior and traces it to an incorrect reuse of \texttt{tree\_draft\_ids} after verification. This bug corrupts the following embedding lookup, which lowers token acceptance and reduces throughput. Removing the incorrect assignment produces the largest gain. AutoSOTA then makes two further runtime improvements. It replaces a Python-side terminal-token check with a GPU-side \texttt{torch.isin} implementation, and it removes repeated CPU--GPU synchronization in the decoder by introducing a precomputed padding path. A final adjustment to the speculative decoding configuration yields a small additional gain. These changes increase throughput from 674.58 to 765.27 tok/s, a \textbf{13.44\%} improvement over the baseline. This case shows that AutoSOTA can find hidden performance bottlenecks in real LLM inference pipelines and turn code-level diagnosis into substantial efficiency gains.

\begin{figure}
    \centering
    \includegraphics[width=\linewidth]{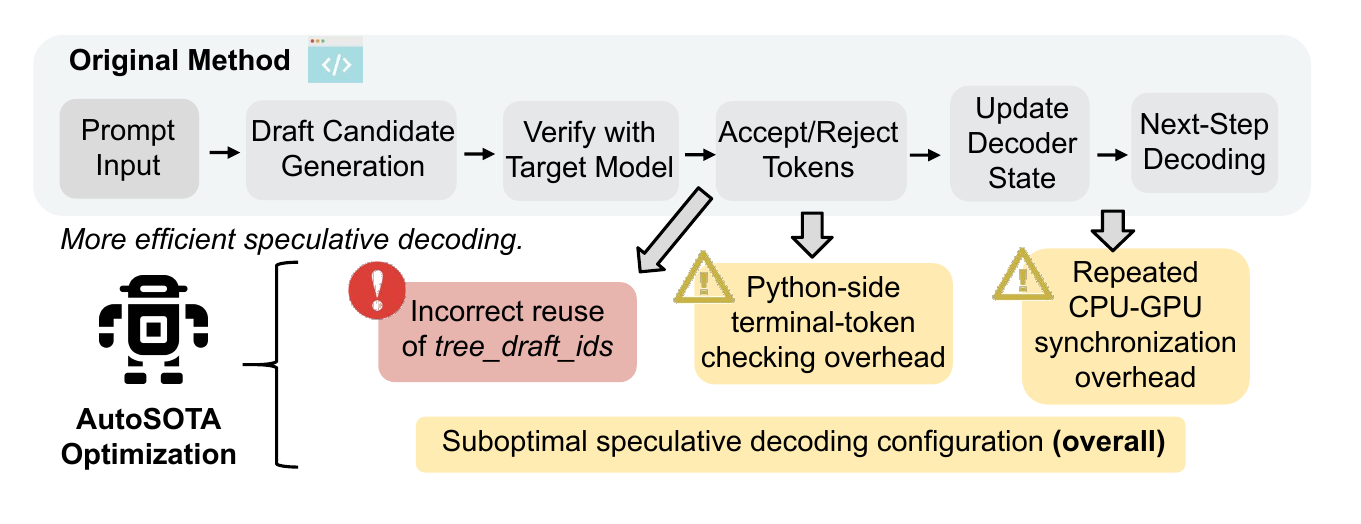}
    \caption{Illustration of the LLM case study.}
\end{figure}

\subsubsection{Case Study for NLP Research.}

We use \textit{Comparing Moral Values in Western English-speaking societies and LLMs with Word Associations} as the NLP case because it provides a clear example of optimization in a graph-based language analysis pipeline. The gain does not come from a larger model or additional supervision. Instead, AutoSOTA improves the target metric by changing how moral information is propagated through the word-association graph. The paper studies moral-value inference with a Graph Moral Network, where moral signals are propagated over a word-association graph. We optimize the primary metric \texttt{correlation\_care}, and the released implementation gives a baseline of 0.4577.

The optimization proceeds in stages. AutoSOTA first replaces raw edge-frequency counts with a triple-\texttt{log1p} transformation before graph normalization. This smooths extreme edge weights and prevents a small number of highly frequent pairs from dominating the graph, which raises the score to 0.4749. AutoSOTA then improves the result further by tuning the propagation strength. In the final stage, it replaces the original single-scale propagation rule with a multi-scale design. Instead of using a single resolvent operator $(I-\alpha T)^{-1}$, the optimized pipeline composes multiple resolvent operators under a decreasing $\alpha$ schedule, with the six-step setting $\alpha=[0.84, 0.82, 0.80, 0.68, 0.49, 0.25]$. This turns the original rule into a multi-scale graph filter that captures semantic structure at different diffusion depths. These changes improve \texttt{correlation\_care} from 0.4577 to 0.4970, a relative gain of \textbf{8.59\%}. The steady progression from edge-weight smoothing to multi-scale propagation also shows that the gain is not incidental. It comes from \textbf{redesigning the mechanism} itself, without a larger model, extra supervision, or task reformulation.

\begin{figure}
    \centering
    \includegraphics[width=\linewidth]{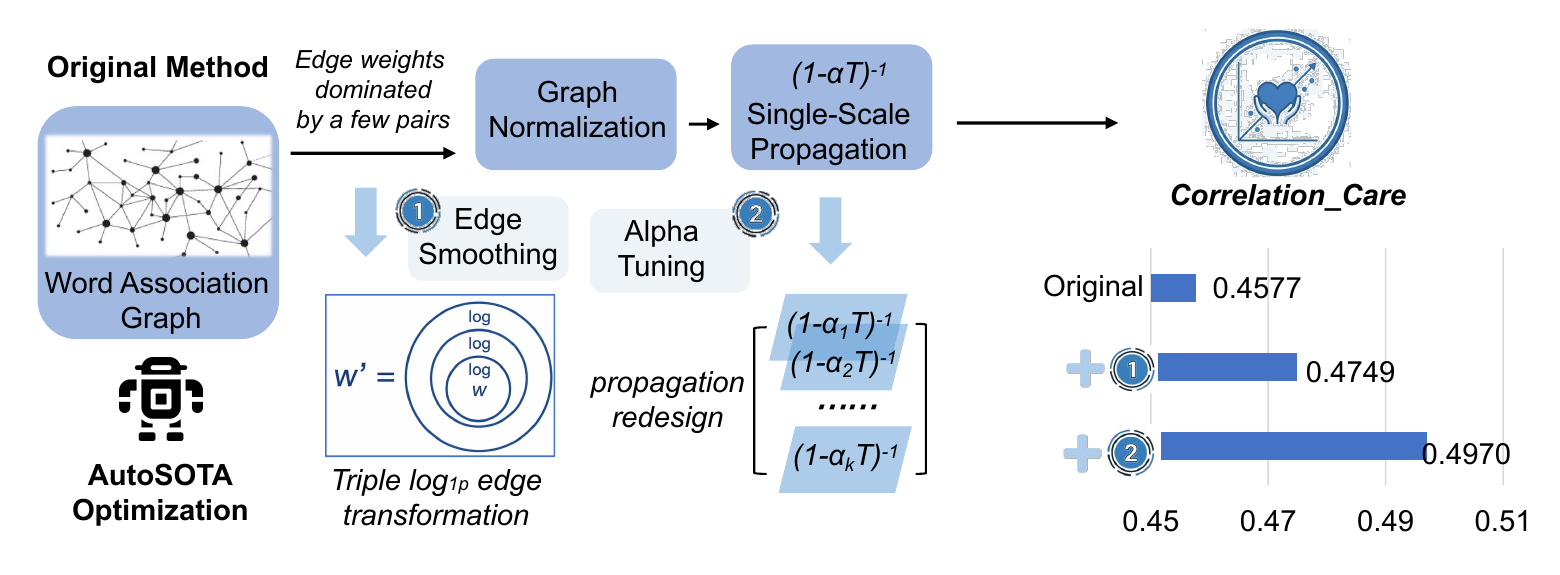}
    \caption{Illustration of the NLP case study.}
\end{figure}

\subsubsection{Case Study for Biology Research.}

We select \textit{Predicting mutational effects on protein binding from folding energy} as the Biology case because it illustrates AutoSOTA's ability to look beyond the core neural architecture and exploit orthogonal, physics-based signals through data-level integration. The task involves predicting the impact of mutations on protein-protein binding affinity ($\Delta\Delta G$). The primary evaluation metric is the per-interface Pearson correlation. Starting from the default evaluation pipeline, AutoSOTA reproduces an initial baseline correlation of 0.4897.

The optimization strategy in this case does not rely on tuning the neural network's internal hyperparameters. Instead, AutoSOTA identifies that the repository already contains pre-computed prediction files from FoldX, a classical physics-based energy calculator. Recognizing that these physics-based predictions capture structural packing effects that are highly complementary to the ML-based model's (StaB-ddG) learned representations, AutoSOTA formulates a hybrid ensemble strategy. Crucially, before fusing the predictions, AutoSOTA autonomously discovers and resolves a subtle data-alignment bug: a spacing inconsistency in the mutation string formats that previously limited the data intersection to only 63\%. After fixing this string formatting to achieve a 100\% row match, AutoSOTA systematically sweeps the blending weights and identifies an optimal fusion ratio of 60\% StaB-ddG and 40\% FoldX.

This ensemble combination immediately increases the per-interface Pearson correlation from 0.4897 to 0.5655, achieving a \textbf{15.48\%} relative improvement in a single iteration. It also drives consistent downstream improvements across other metrics, including an 11.0\% gain in overall Pearson correlation and a 7.7\% reduction in per-structure RMSE. This case demonstrates AutoSOTA's "researcher-like" intuition in computational biology tasks. It proves that the agent can autonomously diagnose dirty data alignments, intelligently leverage existing repository artifacts, and seamlessly bridge data-driven deep learning with traditional physics-based methodologies to shatter single-model performance ceilings.

\begin{figure}
    \centering
    \includegraphics[width=\linewidth]{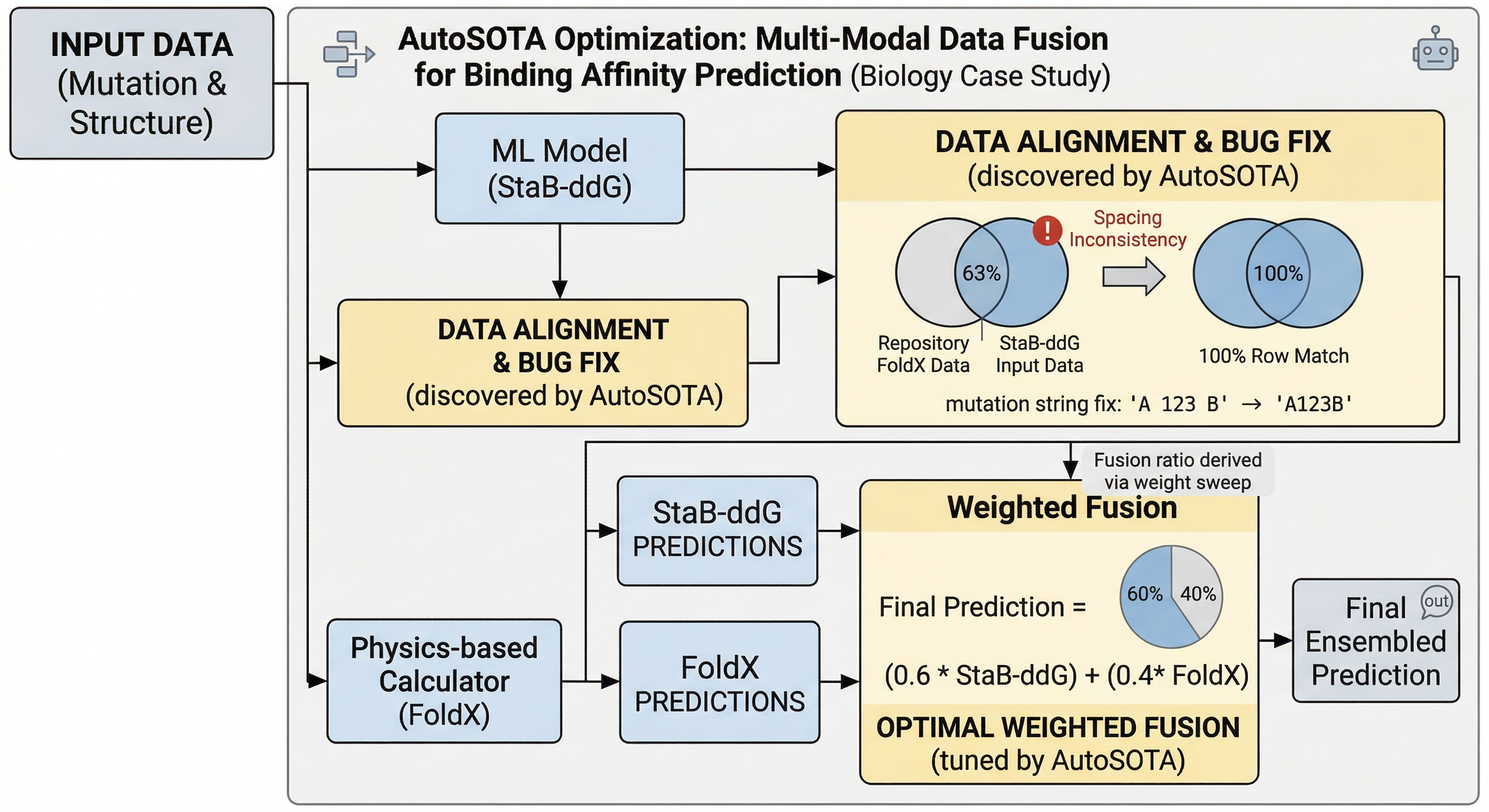}
    \caption{Illustration of the Biology case study.}
\end{figure}

\subsubsection{Case Study for CV Research.}

We select \textit{Reasoning as Representation: Rethinking Visual Reinforcement Learning in Image Quality Assessment} as the representative Computer Vision (CV) case because it demonstrates AutoSOTA's ability to recognize and restructure representation bottlenecks in Vision Transformer (ViT) architectures. The paper tackles the task of Image Quality Assessment (IQA), where both global compositional aesthetics and local artifact distortions are critical for accurate evaluation. The primary evaluation metric is the Pearson Linear Correlation Coefficient (PLCC). Starting from the released implementation, AutoSOTA establishes a reproduced baseline PLCC of 0.7803.

The optimization focuses on the feature aggregation stage preceding the final quality prediction head. In the original implementation, the model relied on a single pooling path to compress visual features. AutoSOTA identifies this as a potential information bottleneck and redesigns the token fusion mechanism to better capture multi-scale quality degradation. Instead of discarding spatial patch details, AutoSOTA introduces a three-way feature blending strategy. It explicitly concatenates the global \texttt{CLS} token with both the average-pooled (\texttt{patch\_mean}) and max-pooled (\texttt{patch\_max}) statistics of the spatial patch tokens, while simultaneously introducing and tuning a blending weight (\texttt{alpha\_cls}). This architectural modification ensures that both holistic global context and salient local evidence jointly contribute to the final scoring mechanism. These targeted structural changes improve the PLCC from 0.7803 to 0.8012, an improvement of \textbf{2.68\%} over the baseline. This case highlights AutoSOTA's deep empirical understanding of modern visual architectures, proving its capacity to autonomously apply classic feature-stitching paradigms to extract richer representations without necessitating larger backbone models or additional training data.

\begin{figure}
    \centering
    \includegraphics[width=\linewidth]{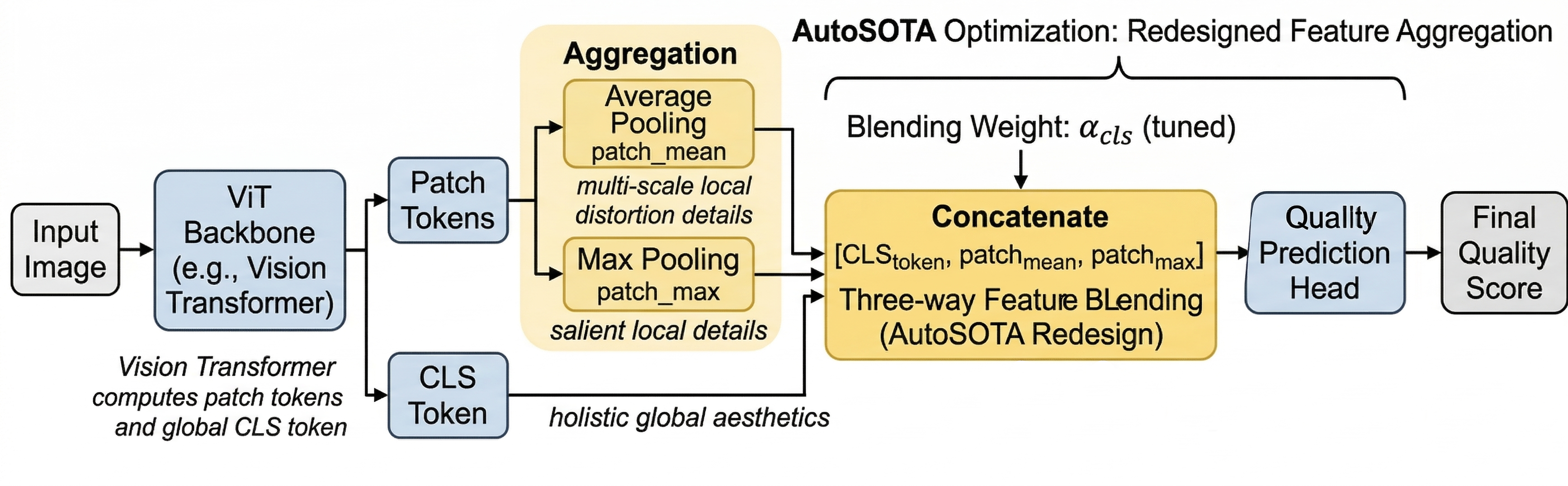}
    \caption{Illustration of the CV case study.}
\end{figure}

\subsubsection{Case Study for Time Series Research.}

\begin{figure}[!b]
    \centering
    \includegraphics[width=1\linewidth]{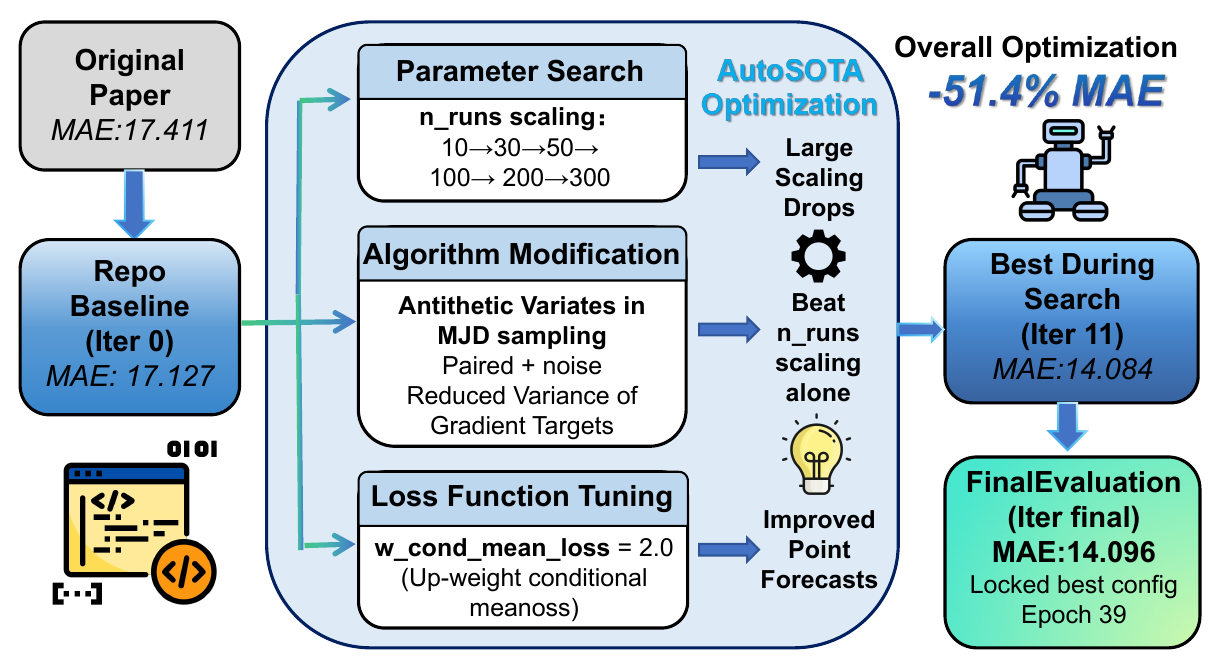}
    \caption{Illustration of the time series case study.}
\end{figure}

We select "Neural MJD: Neural Non-Stationary Merton Jump Diffusion for Time Series Prediction" to demonstrate AutoSOTA's capacity for deep, algorithm-level intervention in complex stochastic models. The task focuses on predicting highly volatile financial data (S\&P 500) using stochastic differential equations (SDEs), evaluated by test Mean Absolute Error (MAE). Starting from the released configuration, AutoSOTA establishes a reproduced baseline of 17.127, which inherently outperforms the paper's reported 17.411. The optimization targets the high-variance bottleneck in the model's Monte Carlo sampling of jump-diffusion paths. First, AutoSOTA dynamically scales the sampling size (n\_runs), identifying 300 as the optimal sweet spot within a fixed compute budget while rejecting higher values that degrade convergence. Second, it introduces antithetic variates to the MJD sampling, generating paired positive and negative noise. This advanced statistical technique effectively reduces gradient target variance and outperforms pure scaling. Finally, AutoSOTA reallocates the conditional mean loss weight from 1.0 to 2.0, penalizing point forecast inaccuracies without overfitting.

These targeted algorithmic refinements drive the MAE down to a locked-in final performance of 14.096 (peaking at 14.084 during search), representing an approximate \textbf{19.0\%} improvement over the original paper's baseline. Crucially, this case highlights AutoSOTA's dynamic optimization and strong self-correction mechanisms under resource constraints. Beyond merely recognizing the compute boundary at 300 sampling runs, the system actively filters out pseudo-optimizations. For instance, AutoSOTA tested an EMA of 0.999 and finer SDE integration steps; while these offered marginal gains in Negative Log-Likelihood (NLL), they failed to translate into MAE improvements and were thus intelligently discarded. By moving beyond surface-level hyperparameter tuning to implement domain-specific variance reduction and robust error filtering, AutoSOTA ensures practical utility and stability for highly volatile time series environments.

\subsubsection{Case Study for Optimization Research.}

We select "MoSES: Multi-Task Vehicle Routing Solver via Mixture of Specialized Experts under State-Decomposable MDP" to demonstrate AutoSOTA's capability in complex, discrete constraint environments. The task focuses on solving the Capacitated Vehicle Routing Problem (CVRP), where the primary metric is the optimality gap percentage. AutoSOTA establishes an initial baseline gap of 0.914\%. To optimize this hybrid neuro-symbolic solver, AutoSOTA executes a multi-layered intervention strategy. First, it modifies the neural architecture's gating mechanism, switching the Low-Rank Adaptation (LoRA) activation function from softplus to a softmax checkpoint to improve expert specialization. Next, recognizing a log-linear scaling trend, AutoSOTA aggressively scales test-time symmetric augmentations from 8 to 128. To circumvent out-of-memory failures during this intensive exploration phase, the system autonomously downscales the batch size from 1000 to 250. Finally, AutoSOTA injects classical Operations Research heuristics for algorithmic post-processing, applying intra-route 2-opt exchanges and highly effective cross-route Or-opt single-node relocations to dynamically move customers between vehicle routes and eliminate final inefficiencies.

This systematic combination of neural adjustments, inference scaling, and deterministic refinement drives the optimality gap down to 0.444\%. This represents a \textbf{51.4\%} relative improvement over the baseline, far exceeding the predefined success target of 0.8957\%. Notably, this aggressive performance gain involves a deliberate computational trade-off, expanding inference time from approximately 3.7 to 103 seconds—a compute reallocation characteristic of modern reasoning-focused search paradigms. Crucially, this case proves AutoSOTA operates effectively as a hybrid optimizer rather than treating the neural network as a mere black box. By seamlessly bridging representation tuning, brute-force state space exploration, and classical algorithmic heuristics, alongside dynamic GPU memory management, AutoSOTA demonstrates an expert-level orchestration of diverse strategies for combinatorial routing problems.

\begin{figure}[!b]
    \centering
    \includegraphics[width=1\linewidth]{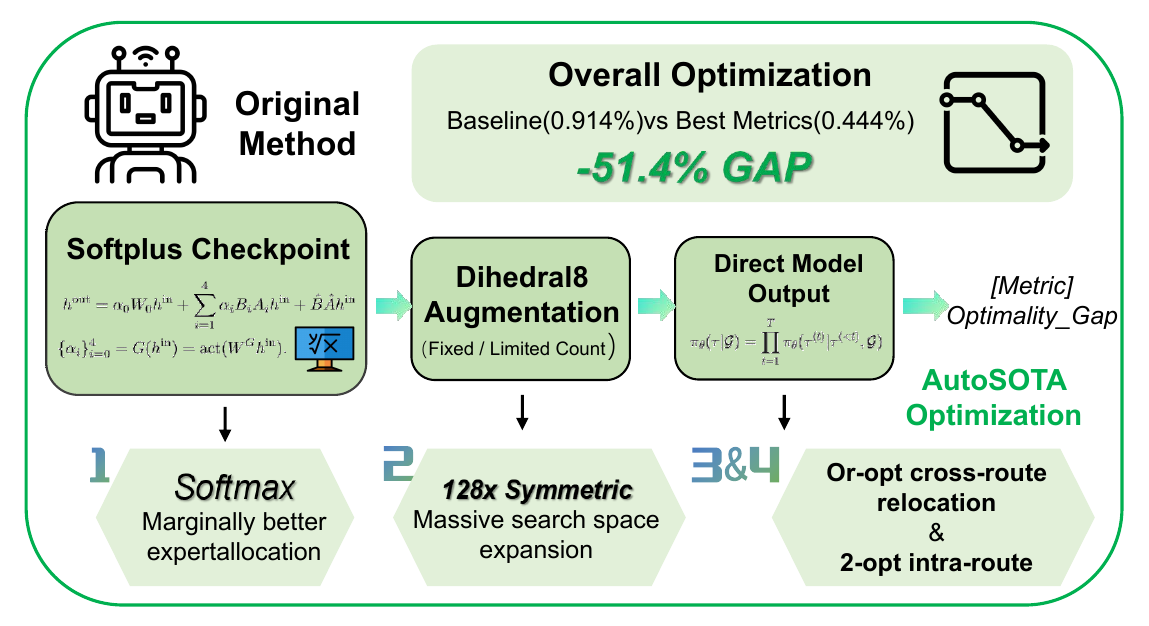}
    \caption{Illustration of the Optimization case study. }
\end{figure}


\section{Related Works}
\label{sec:related_works}

\subsection{Automated Machine Learning (AutoML)}
The foundational goal of automating model design has been extensively explored through Automated Machine Learning (AutoML). Traditional AutoML primarily focuses on Hyperparameter Optimization (HPO) and Neural Architecture Search (NAS). Early NAS approaches utilized reinforcement learning~\cite{zoph2016neural} or evolutionary algorithms~\cite{real2019regularized, shang2024synergy} to navigate discrete spaces of network operations. Subsequent advancements, such as DARTS~\cite{liu2018darts}, introduced differentiable search spaces to dramatically reduce computational overhead. While highly successful in domain-specific applications (e.g., image classification and language modeling), traditional AutoML is inherently limited by its \textit{closed-world assumption}. These systems typically optimize within a rigidly predefined search space (e.g., a fixed set of convolutional kernels or attention heads) and cannot invent novel algorithmic structures, synthesize external knowledge, or debug complex software environments. AutoSOTA transcends these limitations by treating the entire, unrestricted source code of a state-of-the-art repository as an open-ended search space, elevating the paradigm from routine parameter tuning to genuine methodological innovation.

\subsection{LLM-Driven Algorithm Discovery and Code Optimization}
With the advent of advanced LLMs, the focus has shifted from searching over predefined operations to directly generating and optimizing programmatic logic. Systems like SWE-agent~\cite{yang2024swe,shang2024agentsquare,li2026agentswift} and AutoCoder have demonstrated the viability of LLMs in navigating multi-file repositories and resolving complex software engineering issues. Building upon this, researchers have begun leveraging LLMs for scientific and algorithmic discovery. FunSearch~\cite{romera2024mathematical} pairs a pre-trained LLM with an automated evaluator to discover novel mathematical heuristics and algorithms, utilizing a genetic mechanism to evolve Python programs. Similarly, Eureka~\cite{ma2023eureka} employs LLMs to autonomously write and optimize complex reward functions for reinforcement learning tasks, surpassing human-engineered rewards. More recently, evolutionary frameworks like AlphaEvolve~\cite{novikov2025alphaevolve} have attempted to continuously mutate algorithmic code to improve performance. Additionally, open-source initiatives such as AutoResearch~\cite{karpathy2026autoresearch} have explored automating the iterative loop of running experiments, reading logs, and tweaking parameters. However, these code-optimization systems typically operate within highly sanitized, sandbox environments or confine the agent's interventions to a single, bounded code module. They largely bypass the severe engineering bottlenecks of environment initialization, dependency resolution, and long-horizon debugging inherent in unstructured academic codebases. Furthermore, they lack the rigorous supervisory mechanisms required to maintain protocol comparability when modifying complex SOTA repositories---a critical gap that AutoSOTA's \textbf{AgentSupervisor} and red-line constraint system specifically address to prevent spurious empirical gains.

\subsection{Autonomous AI Scientists}
The most ambitious trajectory in recent literature aims to automate the entire scientific research lifecycle, giving rise to ``AI Scientists.'' The AI Scientist~\cite{lu2024ai}, developed by Sakana AI, represents a pioneering effort in fully automated discovery, orchestrating the entire pipeline from idea generation and experimental execution to drafting a complete manuscript. Its second iteration, AI Scientist v2~\cite{yamada2025ai}, demonstrated significant improvements, successfully submitting generated papers to academic venues. Concurrent efforts include DeepMind's AI Co-Scientist~\cite{gottweis2025towards}, which leverages a multi-agent society to propose and collaboratively critique hypotheses in biomedical research, and DeepScientist~\cite{weng2025deepscientist}, which formulates scientific discovery as a Bayesian optimization problem to refine experimental loops. ResearchAgent~\cite{baek2025researchagent} further augments this space by iteratively refining ideas through simulated peer review. While these broad frameworks chart an inspiring path toward fully autonomous scientific discovery, they often operate at a macro level---focusing heavily on conceptual ideation or manuscript generation---without the robust, long-horizon execution grounding required to reliably advance highly competitive empirical baselines. AutoSOTA complements this overarching vision by providing a specialized, fault-tolerant infrastructure dedicated explicitly to the grueling pipeline of SOTA model optimization, ensuring that high-level scientific creativity is physically anchored by verifiable and reproducible code execution.

\section{Conclusion}
\label{sec:conclusion}

In this paper, we introduced AutoSOTA, an autonomous, end-to-end multi-agent ecosystem that seamlessly transitions unstructured academic literature into empirically superior algorithmic implementations. By decomposing the massive cognitive workload of scientific research into a tightly coupled, iterative workflow encompassing Resource Preparation, Experiment Evaluation, Code Optimization, and Reflection \& Ideation, AutoSOTA successfully navigates the complex, long-horizon bottlenecks of open-ended code optimization. Crucially, through the coordinated efforts of eight specialized agents and the strict enforcement of a Red Line System via AgentSupervisor, our framework ensures that all discovered algorithmic innovations remain scientifically rigorous, reproducible, and free from methodological shortcuts. 

Beyond serving as a powerful standalone optimizer, AutoSOTA offers a new blueprint for human-AI collaboration and acts as a foundational infrastructure for the next generation of AI research. By absorbing the grueling, repetitive cycles of environment initialization, baseline reproduction, and empirical tuning, researchers can deploy AutoSOTA to instantly operationalize newly published literature. This frees human cognition from low-level software engineering, allowing scientists to redirect their focus toward higher-level theoretical design, causal reasoning, and disruptive conceptual breakthroughs. Furthermore, within the broader trajectory of ``AI Scientists,'' AutoSOTA bridges a critical execution gap. While existing autonomous scientists primarily excel at macro-level ideation or manuscript generation, AutoSOTA provides the missing empirical grounding---ensuring that AI-driven research is rigorously anchored in reproducible, SOTA-surpassing code execution.

Looking forward, several promising avenues remain to fully realize the potential of automated discovery. Future work will explore extending the automated objective rubric to encompass more complex modalities, such as hardware-in-the-loop systems or physical robotics. Additionally, integrating AutoSOTA's robust execution engine with fully autonomous macro-ideation frameworks, such as OmniScientist, presents an exciting frontier. Such an integration could enable a completely closed-loop paradigm where AI agents not only optimize existing state-of-the-art methods but proactively invent, empirically validate, and publish entirely new algorithmic paradigms.

\newpage

\bibliographystyle{unsrt}
\bibliography{reference}

\end{document}